\documentclass[5p,times]{elsarticle}

\usepackage{amsmath}            
\usepackage{epstopdf}           
\usepackage{flushend}           
\usepackage{hyperref}
 
\usepackage{amssymb}
\usepackage{amsthm}

\usepackage{array}
\newcolumntype{M}[1]{>{\centering\arraybackslash}m{#1}}
\usepackage{math_commands}
\renewcommand{\abs}[1]{\left\vert #1 \right\vert}
\usepackage{xcolor}
\newcommand{\correction}[1]{\textcolor{black}{#1}}

\usepackage[english]{babel}
\newtheorem{theorem}{Theorem}

\newtheorem{lemma}{Lemma}

\usepackage{amsthm}
\newtheorem{remark}{Remark}

\newtheorem{definition}{Definition}

\newtheorem{corollary}{Corollary}[theorem] 

\usepackage{natbib}

\begin{document}

\begin{frontmatter}

\title{Memory Capacity of \textcolor{black}{Recurrent Neural Networks} with Matrix Representation }


  \author[ca]{Animesh Renanse \corref{cor1}\fnref{label2}}
\ead{renanse@iitg.ac.in}

\author[riken]{Alok Sharma} 

\author[last]{Rohitash Chandra  \corref{cor1}\fnref{label2}}
\ead{rohitash.chandra@unsw.edu.au}

\address[ca]{Department of Electronics \& Electrical Engineering, Indian Institute of Technology Guwahati, Assam, India }

\address[riken]{ Laboratory for Medical Science Mathematics, RIKEN Center for Integrative Medical Sciences, Yokohama, Japan  }

\address[last]{ Transitional Artificial Intelligence Research Group, School of Mathematics and Statistics, UNSW Sydney,   NSW 2052, Australia  }

\begin{abstract} 
It is well known that canonical recurrent neural networks (RNNs) face limitations in learning long-term dependencies which have been addressed by memory structures in long short-term memory (LSTM)  networks. Neural Turing machines (NTMs) are novel RNNs that implement the notion of programmable computers with neural network controllers that can learn simple algorithmic tasks. Matrix neural networks feature matrix representation which inherently preserves the spatial structure of data when compared to canonical neural networks that use vector-based representation. One may then argue that neural networks with matrix representations may have the potential to provide better memory capacity.   In this paper, we define and study a probabilistic notion of memory capacity based on Fisher information for matrix-based RNNs. We find bounds on memory capacity for such networks under various hypotheses and compare them with their vector counterparts. In particular, we show that the memory capacity of such networks is bounded by $N^2$ for $N\times N$ state matrix which generalizes the one known for vector networks. We also show and analyze the increase in memory capacity for such networks which is introduced when one exhibits an external state memory, such as  NTMs. Consequently, we construct NTMs with RNN controllers with matrix-based representation of external memory, leading us to introduce Matrix NTMs. We demonstrate the performance of this class of memory networks under certain algorithmic learning tasks such as copying and recall and compare it with Matrix RNNs. We find an improvement in the performance of Matrix NTMs by the addition of external memory, in comparison to Matrix RNNs.

\end{abstract}


\end{frontmatter}

\section{Introduction}
 
Several successful attempts have been made to explain learning and knowledge representation from the perspective of kernel methods  \cite{NIPS2018_8076} to dynamical systems viewpoint of recurrent neural networks (RNNs) \cite{Omlin_Giles1992,Omlin_thonberetal1996,chang2019antisymmetricrnn}. A crucial element in knowledge representation studies of RNNs is that they deal only with finite dimensional vector representations. Thus implementing such neural networks in any architecture would eventually lead to flattening of the input data along any mode, given the input is multi-modal to begin with. It can thus be argued that such a flattening layer can cause loss of structural and spatial information in the input, especially in data arising in natural settings like tri-modal images/sequence of images in longitudinal analysis and higher modal datasets arising in the study of human behavior \cite{10.2307/41217886,6553805}.

The introduction of richer inductive biases \cite{mitchell1980need} in the past decade can be seen through recent deep learning architectures such as convolutional neural networks (CNNs) \cite{krizhevsky2012imagenet}, graph-structured representations \cite{teney2016graphstructured,kipf2019contrastive}, and hierarchy of constituents \cite{shen2018ordered}. 
These methods try to preserve the spatial information by providing fewer constraints on the structure of the representation that the model learns in comparison to a traditional neural network \cite{battaglia2018relational}. \textcolor{black}{A parallel \textcolor{black}{branch of study}, which \textcolor{black}{aims to} provide more generalized representations has been introduced by learning higher-order tensors ($\ge 2$) for multi-modal data to preserve the number of dimensions and the modes at the output \cite{10.1007/978-3-319-70096-0_58,su2018tensorial,AAAI159371}}. This avoids the potential loss of the spatial coherence \cite{bengio2013representation} in the input data, which can be incurred due to the flattening operation of the higher order tensors and instead \textcolor{black}{processes} the multi-modal higher-order tensor input data as it is.

A particularly simple use of learning higher-order tensors as representations is in the field of symmetric positive definite (SPD) matrix (an order 2 tensor) learning, which arises largely in covariance estimation tasks like heteroscedastic multivariate regression \cite{muller1987estimation}. In this task, one wishes to learn fundamentally a matrix representation from the input data, thus several extensions of conventional neural networks with vector representations have been introduced to accept matrix inputs as well, such as learning the SPD representations throughout the layers via Mercer's kernel \cite{taghia2019constructing} to incorporating Riemannian structure of the SPD matrices directly into the layers itself \cite{huang2016riemannian}. The matrix representation in such proposals is generated through the bilinear map of the form,
\begin{equation}
\label{bil-map}
    \mY = \sigmoid\left(\mU^T \mX\mV + \mB\right)
\end{equation}

where $\mU$, $\mV$ and $\mB$ are the parameters to be learned of appropriate shapes \cite{gao2016matrix}. Do et al.~ \cite{do2017learning}, with the help of an aforementioned bilinear map, developed the notion of recurrent matrix networks (RMNs). One of the other benefits of RMNs is the reduction of the number of trainable parameters of the model, as the total parameters depend now linearly on any of the input dimensions. This is in contrast to vector-based conventional neural networks, where the number of parameters grows quadratically with the dimensions of the input vector. 
Matrix neural networks have been successfully used in the past for seemingly difficult real-life tasks like cyclone track prediction \cite{8489077} and in high-frequency time series analysis \cite{8476227}, where the preservation of spatial data amongst the modes of input proved to be beneficiary as these networks gave better accuracy than their vector counterparts.
 
Neural Turing machines (NTMs)\cite{graves2014neural}  are RNN models that combine the fuzzy pattern-matching capabilities of neural networks with the notion of programmable computers.   NTMs feature a neural network controller coupled to external memory that interacts with attention mechanisms.   NTMs that feature long short-term memory (LSTM) \cite{hochreiter1997long} for network controllers can infer simple algorithms that have copying, sorting, and associative recall tasks \cite{graves2014neural}. With their potential benefits, it's natural to investigate properties of the matrix representation as generated by the bilinear mapping (Equation \ref{bil-map}). Out of the various possible ways in which this can be checked (such as convergence guarantee and training behavior etc.), we particularly focus on the asymptotic \textit{memory capacity} of the matrix representation by introducing the dynamical systems viewpoint of generic matrix RNNs \cite{do2017learning}, \textcolor{black}{where we study the evolution of state matrix over time and how much the state at a given time holds information about all previous states.}

In this paper, we study  \textit{memory capacity} using matrix representation as used by matrix RNNs and extend the notion to NTMs. We investigate if the matrix (second-order tensor) representation of data has a better memory capacity than the vector representation in conventional neural networks. Intuitively, one might argue that such matrix representations might always be better than vector representations in terms of memory capacity. However, we show theoretically that it isn't the case under some constraints, even though in general, the capacity might be better than vector representations with a bound that is analogous to the one found for vector representations \cite{Ganguli18970}. Hence, it is vital to discover ways to increase the memory capacity, especially given the memory architectures that have been proposed in the past \cite{graves2014neural,graves2016hybrid,santoro2016meta}. 

In order to provide theoretical evaluation, we take into consideration a generic memory-augmented RNN which simply stores a finite amount of past states. \textcolor{black}{We then use theoretical and simulation study to show that it has greater capacity than a simple matrix recurrent architecture without memory; hence, practically showing the increase in memory capacity that is to be expected by addition of a simple memory architecture.} However, for more practical purposes, we extend the idea of \textit{memory networks} \cite{weston2014memory} to also include the second order tensors, thus introducing the \textit{matrix representation stored neural memory}  for matrix NTM (MatNTM).  We report the results of the simulation of the governing equations of memory capacity of matrix RNNs both with and without external memory and also show results of synthetic algorithmic experiments using MatNTMs.

We present the rest of the paper as follows. In Section  2, we present various definitions for memory capacity in neural networks. In Section 3, the overarching goal is to capture the effect of external state memory dynamics. Motivated by previous discussions, Section 4  presents the MatNTM with the procedure of memory lookup. Section 5 presents a simulation study in order to quantitatively observe the theoretical properties obtained by the theorems in the prior sections.  Section 6 presents experiments and results, which is followed by a discussion   in Section 7.

\section{Background}
\subsection{Memory Capacity in Neural Networks}
\label{MCiNN} 

We note that conventional neural networks follow vector-based representation, and our focus in this paper is matrix-based neural networks. The broader work on the memory capacity and the general capacity of conventional neural networks provides multiple notions of the memory capacity of neural networks. Baldi and Vershynin \cite{NIPS2018_7999} defined $\log_2 \abs{S}$ as the memory capacity of the architecture, where $S$ is the set of all functions that it can realize. It has been shown that for a recurrent neural network with $N$ total neurons, the
memory capacity is proportional to $N^3$ \cite{NIPS2018_7999}. Furthermore, the same definition is used to derive that the capacity of a usual $L$ layered neural network with only threshold activation is upper bounded by a cubic polynomial in size of each layer with bottleneck layers limiting this upper bound \cite{baldi2019capacity}. However, the definition of capacity in this case from a memory standpoint is only \textit{expressive} in the sense that it does not refer to any capability of the network to \textit{remember} the past inputs. A stronger argument has been 
for the definition of \textit{memory capacity} where the existence of a realizable function such that the given architecture can perfectly remember past inputs \cite{NIPS2018_7999}. That is, the largest value of $K$ such that there exists a function $F$ in the space of all functions realizable by the given architecture $S$, so that for a given set of data $x_i$ and label $y_i$, $\{(x_1,y_1),\dots,(x_K,y_K)\}$, it's true that
\[F(x_i) = y_i\;\;\text{for all } \; i = 1,\dots,K.\]
 This definition has been used to develop the memory capacity of multilayered neural networks with threshold and rectifier linear unit (ReLU) activation functions \cite{vershynin2020memory}, where it was shown that the memory capacity is lower bounded by the number of total weights. This definition allows deeper mathematical analysis as visible by the remarkable results in other works \cite{vershynin2020memory}.   However, this definition is not easily portable to higher-order representations and their analysis. We gather that most of the analysis done \cite{vershynin2020memory} depends on the vector representations retrieved from the affine transform $\mW\vx + \vb$, whereas this paper focuses on the bilinear transform $\mU^T\mX\mV+\mB$ to generate matrix representations of input data. Apart from this, the set of functions realized by the bilinear transforms are fundamentally different from affine transformation. The extension of the theory would at the very least need extension of the probabilistic view of the realizable function $F$ to their matrix counterpart, which further makes the situation unmanageable \cite{baldi2019capacity,vershynin2020memory}.

Some of the earlier works on neural networks from a dynamical systems viewpoint provides a possible alternative definition of the memory capacity purely in terms of a statistic. A  discrete-time recurrent system  \cite{White_2004,Ganguli18970} is given by 
\begin{equation}
\label{vector-system}
    \begin{split}
        \vx(n) = \mW\vx(n-1) + \vv s(n) + \vz(n)
    \end{split}
\end{equation}
where $\mW$ is the recurrent connectivity matrix and the time-dependent scalar signal $s(n)$ is passed through this system via feedforward connection $\vv$ which contains an additive noise content $\vz(n)$. Additionally, we can consider an output layer that transforms the state $\vx(n)$ of length $N$ to an output vector $\vy(n)$ of length $M$ such that $y_i(n) = \vx(n)^T\vu_i$ for $i=1,\dots,M$. The objective of the system is to have output $y_j(n)$ equal to the past signal observed $s(n-j)$. Given  this premise in place,  the memory trace $m(k)$ has been defined \cite{White_2004}  as
\begin{equation}
    \label{White-defin}
    \begin{split}
        m(k) = \mathbb{E}\left[ s(n-k) y_k(n) \right]
    \end{split}
\end{equation}
where  expectation is over time. The (asymptotic) memory capacity can be now defined naturally as the following extension of Eq. \ref{White-defin},
\begin{equation}
    \label{White-capacity}
    m_{tot} = \sum_{k=0}^{\infty} m(k).
\end{equation}

Using this definition, we are only shifting our focus on a particular task of past recall, which is orthogonal to our aim of determining the memory capacity of the system itself, invariant of the task it's performing. To achieve this aim, we clearly need to focus more on the memory capacity induced by the state transition of the system.  There has been an extension to a more general notion of memory capacity  \cite{Ganguli18970}, with the trace of the Fisher Information Matrix (FIM) or equivalently the Fisher Memory Matrix (FMM). This depends upon the context of interpretation of the state $\vx(n)$ with the vector of temporal scalar inputs $\vs$, as the parameters of the conditional $p(\vx(n) \vert \vs)$. 

Hence, the Fisher information matrix of $p(\vx(n) \vert \vs)$ measures the memory capacity by the capability of past inputs $s(i)$ for $i\le n$ to perturb the current state $\vx(n)$, which in turn shows the reluctance or permeability of the state $\vx(n)$ to change under a new $\vs$. Note that such a measure does not depend explicitly on the structure of the state $\vx(n)$ as that information is portrayed by $p(\vx(n)\vert \vs)$. Thus, as long as we are able to determine the conditional for a given type of state and discrete dynamics (even for matrix representations shown in Eq. \ref{dyn-sys}), we can use this definition for obtaining a measure of memory capacity. However, its reliance on statistical information about the sensitivity of state rather than an explicit measure of the maximum size of the input data which the architecture can map correctly, makes this definition weaker  
than that given by Vershynin  \cite{vershynin2020memory} for artificial neural networks.  However, due to its invariance on the state structure explicitly, and dependence only on the conditional distribution of the current state given past inputs, it becomes ideal for analysis on a much more diverse set of representations in neural networks.

\subsection{Memory Trace in Matrix Representation of Neural Networks}
\label{MTiMRN}

 In this section, we define the recurrent dynamical system (that denotes a time-dependent evolution of states) with matrix representations and then derive its Fisher Memory Curve. We then define and derive the \textit{memory capacity} \textcolor{black}{of matrix RNNs} under certain constraints, which further motivates the need for external memory augmentation.

We first study the following dynamical system
\begin{equation}
	\label{dyn-sys}
	\begin{split}
		\mX\left( n \right) = f\left( \mU^{T}\mX\left( n-1 \right)\mV + \mW s\left( n \right)  + \rmZ\left( n \right) \right)
	\end{split}
\end{equation}
where $f\left( . \right) $ is a pointwise non-linearity, $s\left( n \right) $ is a time-dependent scalar signal, $\mX\left( t \right)  \in \R^{N\times N}$ is the recurrent state of the network, $\mW \in \R^{N\times N}$ is the feedforward connection for the scalar signal to enter the state, $\mU^{T}$ and $\mV \in \R^{N\times N}$ are the connectivity matrices which transforms current state to the next state and $\rmZ \sim \mathcal{M}\mathcal{N}_{N\times N}(\mathbf{0},\varepsilon_1\mI,\varepsilon_2\mI)$ is the additive matrix Gaussian noise with row-wise covariance being $\varepsilon_1\mI$ and column-wise covariance being $\varepsilon_2\mI$ \cite{gupta1999matrix}.

A recurrent matrix representation system similar to Eq. \ref{dyn-sys} has been developed recently \cite{do2017learning} for dealing with temporal matrix sequences and successfully used for sequence-to-sequence learning tasks \cite{sutskever2014sequence}. In particular, the authors used the following bilinear transformation
\begin{equation}
\label{DoEq}
    \mH_t = f\left( \mU_1^T \mX_t \mV_1 + \mU^T_2 \mH_{t-1}\mV_2 + \mB\right)
\end{equation}
where it's clear on comparison with Eq. \ref{dyn-sys} the main difference that Eq. \ref{dyn-sys} deals with scalar input $s(n)$ whereas Eq. \ref{DoEq} with more general matrix inputs $\mX_t$. \textcolor{black}{However,  since our task is to elucidate how much information is stored in the current state about the input signal appearing in the past, we continue with Eq. \ref{dyn-sys}.} 

\subsection{The Fisher memory curve}
\label{The-FMC}


We adopt the framework given by Ganguli et. al \cite{Ganguli18970} to develop the performance measure to formalize the efficiency of 
RNN state matrix $\mX$ to encode the past input signals. Since the information contained in the input signal $s\left( t \right) $ is transferred to the recurrent state via the connection matrix $\mW$, after large enough  $t$, the past signal history induces a conditional distribution on network state  $\mX$; denoted as $p\left( \mX\left( n \right)  \mid \vs \right) $, where $\vs$ is the vector of all past input signals, with $\vs_k = s\left( n-k \right) $.

We focus on the amount of information that is contained in $\mX\left( n \right) $ of an input signal which appeared $k$ time steps in the past, i.e. $\vs_k$. We would thus need to know how much the conditional distribution \textcolor{black}{of} network state $p\left( \mX\left( n \right)  \mid \vs \right) $ changes with change in $\vs$. We note that the Kullback–Leibler (KL) divergence \cite{kullback1951information,Ganguli18970,kullback1997information,csiszar2004information} which is also known as relative entropy is a measure of how one probability distribution differs from another probability distribution which is useful in comparing statistical models.   The KL-divergence  for a small change $\delta\vs$, between  $p\left( \mX\left( n \right)  \vert \vs \right) $ and $p\left( \mX\left( n \right)  \vert \vs + \delta\vs\right) $ can be shown via second-order Taylor series expansion to be approximately\footnote{Note that this approximation becomes an equality if the mean $\mM(\vs) = \mathbb{E}\left[\mX(n)\vert \vs \right]$ is only linearly dependent on $\vs$, as is the case in Eq. \ref{mean-cov-state-sol} \textcolor{black}{below} \cite{martens2020new}.} equal to $\frac{1}{2}\delta\vs^{T}\mJ \left( \vs \right) \delta\vs$; where $\mJ\left( \vs \right) $ is the Fisher Memory Matrix whose element $\mJ_{i,j}$ identifies the sensitivity of $p\left( \mX\left( n \right)  \mid \vs \right) $ to interference between $\vs_i$ and $\vs_j$ which are the input signals appearing $i$ and $j$ timesteps in the past respectively and is given as 

\begin{equation}
	\label{fmm}
	\begin{split}
		\mJ_{i,j}\left( \vs \right)  = \E_{p\left( \mX\left( n \right)  \mid \vs \right) } \left[ - \frac{\partial ^2}{\partial \vs_i \partial \vs_j} \log p\left( \mX\left( n \right)  \mid \vs \right) \right]
	\end{split}
\end{equation}

which is parameterized by all the past inputs $\vs$. It's trivial to see now that the diagonal elements of Fisher Memory Matrix (FMM) $\mJ_{i,i}$ represent the Fisher information that the current state $\mX\left( n \right) $ contains about the input signal that appeared $i$ time steps in the past, i.e.  $\vs_i$. \textcolor{black}{Hence, $\mJ_{i,i}$ can be considered as the decay of the information about signal $\vs_i$ which was presented $i$ time steps into the past to the network} giving the name, Fisher Memory Curve (FMC). Deriving the FMC for Eq. \ref{dyn-sys} will thus provide us with a quantitative measure for the memory decay (and thus, the total capacity) in its dynamics. We first state and prove the FMM for the dynamical system defined by Eq. \ref{dyn-sys} as the theorem which follows and then use it to derive its FMC and then the consequent bounds on memory capacity under various assumptions.
\textcolor{black}{\begin{remark}
In the following results, we assume the absence of any non-linearity $f(\cdot)$ in the state (Eq. \ref{dyn-sys}). This assumption will be reversed in Section \ref{S-EoFDR}, when we consider what happens when given saturating non-linearities.
\end{remark}}

\begin{theorem}\label{T-2}\textit{(FMM for system in Eq. \ref{dyn-sys})  
Consider the recurrent system defined by \ref{dyn-sys} \textcolor{black}{where $f(\cdot) = \text{id}(\cdot)$}, then the Fisher Memory Matrix defined in \ref{fmm} for such a system is
\begin{equation}
	\label{fmm-simpl-1}
	\begin{split}
		\mJ_{i,j}\left( \vs \right) = \Tr\left( \mathbf{\Sigma}^{-1} \frac{\partial \mM\left( \vs \right)^{T} }{\partial \vs_i} \mathbf{\Psi}^{-1}\frac{\partial \mM\left( \vs \right) }{\partial \vs_j} \right) 
	\end{split}
\end{equation}
where,
\begin{equation}
	\label{mean-cov-state-sol}
	\begin{split}
		\mM\left( \vs \right) &=  \sum_{k=0}^{\infty} \mU^{kT}\mW\vs_k\mV^{k} \\	\mathbf{\Psi} &= \varepsilon_1\sum_{k=0}^{\infty} \mU^{kT}\mU^{k}\\
		\mathbf{\Sigma} &= \varepsilon_2\sum_{k=0}^{\infty} \mV^{kT}\mV^{k}.
	\end{split}
\end{equation}
 }\end{theorem}
 \begin{proof}
 The assumption which focuses only on linear dynamics simplifies the analysis by a big margin as we will see in later sections. Given  this  assumption of linear dynamics,
the canonical solution of the linear form of Eq. \ref{dyn-sys} can be derived easily, and the final result is stated below:
\begin{equation}
	\label{state-sol}
	\begin{split}
\mX\left( n \right) = \sum_{k=0}^{\infty} \mU^{kT}\mW\vs_k\mV^{k} + \sum_{k=0}^{\infty} \mU^{kT}\rmZ\left( n-k \right) \mV^{k}.	
	\end{split}
\end{equation}
\newline
Since the noise is an additive Gaussian matrix $\rmZ$, this implies that $p\left( \mX\left( n \right)\vert\vs \right) $ will also be a Gaussian distribution; though with different mean and covariance matrices, where there are two covariance matrices of size $N\times N$, one for rows as random vectors and another for columns. Since the noise entering the system at different times are also independent, we can write the mean matrix $\mM$, the row-wise covariance matrix $\mathbf{\Psi}$, and the column-wise covariance matrix $\mathbf{\Sigma}$ for $\mX\left( n \right) $ as shown in Eq. \ref{mean-cov-state-sol}. 
Note that the mean matrix $\mM$ is parameterized linearly by input signal $\vs$ while the covariance matrices are independent of $\vs$.\\

We have seen in  Section \ref{The-FMC}   
that the KL-divergence can be approximated via quadratic Taylor series approximation. Calculating the KL-divergence of $p\left( \mX\left( n \right)\vert\vs \right) $ will thus allow us to write the FMM in terms of $\mM$, $\mathbf{\Psi}$ and $\mathbf{\Sigma}$. Hence, we now state the form of KL-divergence between two matrix Gaussian distributions as the following lemma,
\newline\newline
\begin{lemma}\label{L-3}($\KL$ for $\mathcal{MN}$) The KL-divergence between $p\left( \rmX_1 \right) $ and $p\left( \rmX_2 \right) $ where $\rmX_1 \sim \mathcal{MN}_{n\times p}\left(\mM_1, \mathbf{\Sigma}_1, \mathbf{\Psi}_1\right)$ and\\ $\rmX_2 \sim \mathcal{MN}_{n\times p}\left(\mM_2, \mathbf{\Sigma}_2, \mathbf{\Psi}_2\right)$ is given by
\begin{equation*}
	\begin{split}
		\KL \left( p\left( \rmX_1 \right) \Vert p\left( \rmX_2 \right)  \right) = &\frac{1}{2} \Biggl[ \log \frac{\vert \mathbf{\Sigma}_2\vert^{p} \vert \mathbf{\Psi}_2\vert^{n}}{\vert \mathbf{\Sigma}_1\vert^{p}\vert\mathbf{\Psi}_1 \vert^{n}}  -np + \Tr\left( \mathbf{\Psi}_2^{-1}\mathbf{\Psi}_1 \right) \Tr\left( \mathbf{\Sigma}_2^{-1}\mathbf{\Sigma}_1 \right)\\& + \Tr\left( \mathbf{\Sigma}_2^{-1}\left( \mM_2 - \mM_1 \right) ^{T} \mathbf{\Psi}^{-1}_2\left( \mM_2 - \mM_1 \right)  \right)  \Biggl]  
	\end{split}
\end{equation*}
where $\mM_1$, $\mathbf{\Psi}_1$, $\mathbf{\Sigma}_1$ and $\mM_2$, $\mathbf{\Psi}_2$, $\mathbf{\Sigma}_2$ are the mean and covariance matrices for $\rmX_1$ and $\rmX_2$, respectively.
\end{lemma}

We provide proof for the above lemma  in Appendix \ref{Appendix-A1}. Notice that from Eq. \ref{mean-cov-state-sol}, the independence of $\mathbf{\Psi}$ and $\mathbf{\Sigma}$ from signal history $\vs_k$, thus the required $\KL \left( p\left( \mX\left( n \right) \vert \vs^{1} \right) \Vert p\left( \mX\left( n \right) \vert \vs^{2} \right)  \right) $ for different histories $\vs^{1}$ and $\vs^{2}$ simplifies to: 
\begin{equation}
	\label{KL-div-diff-s}
	\begin{split}
		&\KL \left( p\left( \mX\left( n \right) \vert \vs^{1} \right) \Vert p\left( \mX\left( n \right) \vert \vs^{2} \right)  \right)\\ &= \frac{1}{2}\left[ \Tr\left( \mathbf{\Sigma}^{-1} \left( \mM_2-\mM_1 \right) ^{T}\mathbf{\Psi}^{-1}\left( \mM_2 - \mM_1 \right)   \right)  \right]. 
	\end{split}
\end{equation}
Thus, for the small change in input signal $\vs_1 = \vs$ by $\delta \vs$ (hence $\vs_2 = \vs + \delta\vs$), the \ref{KL-div-diff-s} would be approximately equal to $\frac{1}{2}\delta \vs^{T} \mJ\left( \vs \right) \delta \vs$ as discussed earlier which would directly yield us the FMM as defined in Eq. \ref{fmm-simpl-1}, hence completing the proof.
 \end{proof}

 \textcolor{black}{We can now derive the FMC for linear version of Eq. \ref{dyn-sys} quite easily as  follows.} \begin{corollary}\label{C-4} \textit{(FMC for system in Eq. \ref{dyn-sys}) The Fisher Memory Curve for the recurrent system \ref{dyn-sys} is
 \begin{equation}
	\label{fc}
	\begin{split}
		\mJ\left( i \right) = \mJ_{i,i} &= \Tr\left( \mathbf{\Sigma}^{-1}\mV^{iT} \mW^{T}\mU^{i}\mathbf{\Psi}^{-1}\mU^{iT}\mW\mV^{i} \right).
	\end{split}
\end{equation}
 } \end{corollary}

 \begin{proof}
Since $\mJ_{i,i}$ is the $(i,i)$-diagonal entry of the FMM $\mJ_{i,j}$ as in Theorem \ref{T-2} and since the derivative of $\mM(\vs)$ w.r.t. $\vs_i$ is trivial to compute, we get the desired result.
 \end{proof}
\begin{remark}\label{R-5}
To obtain an even simpler version of the FMM of Eq. \ref{fmm-simpl-1}, one can further assume that the type of dependency of the mean $\mM\left( \vs \right) $ is linear with $\vs$ as is the case in \ref{mean-cov-state-sol}, which directly implies that Eq. \ref{fmm-simpl-1} is independent of $\vs$, leading us to the following final form of FMM of Eq. \ref{fmm-simpl-1}:
\begin{equation}
	\label{fmm-simplest}
	\begin{split}
		\mJ_{i,j}  &=  \Tr\left( \mathbf{\Sigma}^{-1}\mV^{iT}\mW^{T}\mU^{i} \mathbf{\Psi}^{-1} \mU^{jT}\mW \mV^{j} \right).
	\end{split}
\end{equation}
\end{remark}
One should note that the form in Eq. \ref{fmm-simplest} is not as trivial to analyze as the $\mJ_{i,j} $ has been  given  \cite{Ganguli18970} for linear dynamical system with vector representations. The FMM equivalent for Eq. \ref{fmm-simplest} for such (vector representation) networks takes a very simple form as 
\begin{equation}
	\label{fmm-vec-rep}
	\begin{split}
	\mJ_{i,j} = \vv^{T} \mW^{iT}\mC_n^{-1}\mW^{j}\vv
	\end{split}
\end{equation}
where $\vv$ is the feedforward connection from input to state,  $\mW$ is the recurrent connectivity matrix and  $\mC_n$ is the covariance matrix for the state given by  $\mC_n = \sum_{k=0}^{\infty} \mW^{k}\mW^{kT}$. Extensive analysis is possible with Eq.  \ref{fmm-vec-rep} as done in earlier works \cite{Ganguli18970} and further extension of the theory to other neural architectures such as \textit{echo state networks} \cite{tino2017fisher}. However, there's been little to no work in other domains of representations and the analysis of their capabilities from a memory standpoint. The break in symmetry of Eq. \ref{fmm-simplest} is clear in comparison to Eq. \ref{fmm-vec-rep}, which is the main cause of complicated interaction terms between the recurrent and the feedforward connections which in turn decreases the capacity as we shall see in the following sections.

We are mostly concerned with the information retained in $p\left( \mX\left( n \right) \vert \vs \right) $ about a signal $\vs_k$ entering the network $k$ time steps into the past. Considering FMM given in Eq. \ref{fmm-simplest}, this is expressed by the diagonal elements of $\mJ$ which signifies the Fisher information that the state $\mX\left( n \right) $ retains about a pulse entering the network $k$ time steps in the past. Thus, using this concept, one can define the memory decay of previous inputs by considering the set  $\left\{ \mJ\left( i \right) = \mJ_{i,i} \vert\; 0 \le i<\infty  \right\} $ of all diagonal elements of $\mJ$, hence the FMC as described in Eq. \ref{fc}.

\subsection{The Memory Capacity}\label{Sec-TMC}

The FMC shown in Eq. \ref{fc} identifies the decay of information in the state $\mX\left( n \right) $ about the past signals. However, in order to capture the memory capacity of the system, we would need to measure the amount of information remaining in $\mX\left( n \right)$ for all the previous inputs as this will tell us the exact amount of Fisher information that the system has encoded about all the prior inputs. This can be represented by summing over all $i$ in Eq.  \ref{fc}, thus yielding us the following definition of the memory capacity.
\begin{definition}\label{D-6}
(Memory Capacity) The asymptotic memory capacity of a recurrent system is defined as the sum of the Fisher information stored for all of the past input signals in the current state of the system
\begin{equation}
	 \label{J-tot-sum-form}
	\begin{split}
		\mJ_{tot} &=  \sum_{i=0}^{\infty} \mJ\left( i \right).
	\end{split}
\end{equation}
\end{definition}
Given this definition, the memory capacity of the matrix representation of RNN as defined in Eq. \ref{dyn-sys} with no non-linearity ($f=\text{id}$) is
\begin{equation}
\label{J-tot-mat-rep}
\mJ_{tot} = \sum_{i=0}^{\infty} \Tr\left( \mathbf{\Sigma}^{-1}\mV^{iT}\mW^{T}\mU^{i}\mathbf{\Psi}^{-1}\mU^{iT}\mW\mV^{i} \right).
\end{equation}
This definition of memory capacity is general but doesn't give much insight into the limit or bounds of the capacity directly. Motivated by the study of vector representation on the similar assumptions given in \cite{Ganguli18970},  
we now discuss two cases. Firstly, we consider the case when the recurrent connectivity matrices are assumed to be normal, and secondly, the case when they are not.  We discuss the latter with a reformed view of Eq. \ref{fc}.


\subsubsection{Capacity of Normal Networks}\label{Sec-CNN}

The form of $\mJ\left( i \right) $ in  Eq. $\ref{fc}$ doesn't allow much room for direct analysis, thus, we make certain assumptions to dissect it further. One important assumption that we will continue to deal with is that of normal recurrent connectivity i.e. matrices $\mU^{T}$ and $\mV$ are assumed to be normal, that is they commute with their transpose. This assumption simplifies the FMC and provides insights into the interaction between the connectivity matrices. Hence, we can write  Eq. \ref{fc} as the eigendecomposition of $\mU$ and $\mV$ 
\begin{equation}
	\label{J-tot-eigdecomp}
	\begin{split} 
\mJ(i) &= \frac{1}{\varepsilon_1\varepsilon_2} \Tr\left(  \left( \mI - \mathbf{\Lambda}^2_V \right) \mathbf{\Lambda}_V^{i}\mB^{H}\mathbf{\Lambda}_U^{i}\left( \mI - \mathbf{\Lambda}_U^{2} \right) \mathbf{\Lambda}_U^{i}\mB\mathbf{\Lambda}_V^{i} \right)
	\end{split}
\end{equation}
where $\mE_V$,  $\mathbf{\Lambda}_V$ and $\mE_U$,  $\mathbf{\Lambda}_U$ are the corresponding orthogonal eigenvector and diagonal eigenvalue matrices for $\mV$ and $\mU$ respectively and for clarity, we define $\mB := \mE_U^{H}\mW\mE_V$. The form in Eq. \ref{J-tot-eigdecomp} can be achieved by substituting  Eq. \ref{mean-cov-state-sol}  
in Eq. \ref{fc} and taking all matrices inside the inverse introduced by $\mathbf{\Sigma}$ and  $\mathbf{\Psi}$.

Deriving the capacity $\mJ_{tot}$ is now straightforward from Eq.  \ref{J-tot-eigdecomp}, and the calculations   \textcolor{black}{are shown as follows}.  
\begin{theorem}\label{T-6}
 (Capacity of Normal Matrix Networks) Given the dynamical system \ref{dyn-sys} and that both the connection matrices $\mU$ and $\mV$ are normal, the memory capacity of the system is given as the following.
\begin{equation}
	\label{J-tot-main-final}
	\begin{split}
		\mJ_{tot} &= \frac{1}{\varepsilon_1\varepsilon_2} \sum_{j=1}^{N} \sum_{k=1}^{N} 	\frac{\left( 1-\abs{\lambda_{V_k}}^2 \right) \left( 1-\abs{\lambda_{U_j}}^2 \right) }{1-\abs{\lambda_{V_k}}^2\abs{\lambda_{U_j}}^2} |b_{jk}|^2
	\end{split}
\end{equation}
where $\lambda_{V_k}$,  $\lambda_{U_j}$ are the respective eigenvalues of $\mV$ and $\mU$ and $b_{jk}$ is the $(j,k)^{th}$ element of $\mB$.
\end{theorem}
\begin{proof}
To derive the form of $\mJ_{tot}$ as in Eq. \ref{J-tot-main-final} from  $\mJ\left( i \right) $ (Eq. \ref{J-tot-eigdecomp}), we would first need to notice that the sum over Trace is just Trace of the sum, enabling us to write,
\begin{equation}
	\label{J-tot-ed-sum}
	\begin{split}
		\mJ_{tot} &=  \frac{1}{\varepsilon_1\varepsilon_2} \Tr\left( \sum_{i=0}^{\infty} \left( \mI - \mathbf{\Lambda}_V^{2}\right) \mathbf{\Lambda}_V^{i} \mB^{H}\mathbf{\Lambda}_U^{2i}\left( \mI - \mathbf{\Lambda}_U^{2}\right)\mB \mathbf{\Lambda}_V^{i}  \right).
	\end{split}
\end{equation}
Since the $\mJ_{tot}$ above includes $\mB$ and its transpose in between the diagonal eigenvalue matrices, thus there will be off-diagonal interaction terms in between the eigenvalues, which would complicate the sum. However, one can note that we can write a matrix equation of the form $\mD_\vx \mB^{H} \mD_\vy \mB \mD_\vz$ where $\mD_\vx$, $\mD_\vy$ and  $
\mD_\vz$ are diagonal matrices formed by column vectors  $\vx$,  $\vy$ and  $\vz$ respectively as follows
\begin{equation*}
	\begin{split}
		\mD_\vx \mB^{H}\mD_\vy \mB \mD_\vz  &=  \sum_{j=1}^{N} \vy_j \left( \vx \odot \vb_j^{H}  \right)   \left( \vz^{H} \odot \vb_j \right) \\
		&= \sum_{j=1}^{N} \vy_j \left( \vx \vz^{H} \right)\odot\left( \vb_j^{H}\vb_j \right) 
	\end{split}
\end{equation*}
where $\vb_j$ is the $j^{th}$ row of $\mB$. Hence, we can represent \ref{J-tot-ed-sum} by letting $\vx^{i} =  \text{vec}\left(\left( \mI - \mathbf{\Lambda}_V^2 \right) \mathbf{\Lambda}_V^{i}\right) $, $\vy^{i} = \text{vec}\left( \left( \mI - \mathbf{\Lambda}_U^2 \right) \mathbf{\Lambda}_U^{2i} \right) $ and $\vz^{i} = \text{vec}\left( \mathbf{\Lambda}_V^{i} \right) $
\begin{equation}
	\begin{split}
	\mJ_{tot} &= \frac{1}{\varepsilon_1\varepsilon_2} \Tr\left( \sum_{i=0}^{\infty} \sum_{j=1}^{N} \vy_j^{i} \left( \vx^{i} (\vz^{i})^{H} \right)\odot\left( \vb_j^{H}\vb_j \right) \right)\\
		&= \frac{1}{\varepsilon_1\varepsilon_2}\Tr\left( \sum_{j=1}^{N} \sum_{i=0}^{\infty} \vy_j^{i} \left( \vx^{i} (\vz^{i})^{H} \right)\odot\left( \vb_j^{H}\vb_j \right) \right)\\
		&= \frac{1}{\varepsilon_1\varepsilon_2}\Tr\left( \sum_{j=1}^{N}  \underbrace{\left(\sum_{i=0}^{\infty}\vy_j^{i}\left( \vx^{i}\left( \vz^{i} \right) ^{H} \right)\right)}_{\text{An $N\times N$ Matrix $\mathbf{\Lambda}_j$}}  \odot \left( \vb_j^{H}\vb_j \right)  \right) 
		\end{split}
\end{equation}
where $\mathbf{\Lambda}_i$ is the matrix of interaction between the eigenvalues of $\mU$ and $\mV$ obtained after taking the infinite geometric sum present in each of its elements, which can be represented as the following matrix
\begin{equation}
	\begin{split}
		\mathbf{\Lambda}_j = \begin{bmatrix} \frac{\left( 1-\abs{\lambda_{V_1}}^2 \right) \left( 1-\abs{\lambda_{U_j}}^2 \right) }{1-\abs{\lambda_{V_1}}^2\abs{\lambda_{U_j}}^2} & \cdots & \frac{\left( 1-\abs{\lambda_{V_1}}^2 \right) \left( 1-\abs{\lambda_{U_j}}^2 \right) }{1-\abs{\lambda_{V_1}}\abs{\lambda_{V_N}}\abs{\lambda_{U_j}}^2}\\
		\vdots & \ddots & \vdots \\
	\frac{\left( 1-\abs{\lambda_{V_N}}^2 \right) \left( 1-\abs{\lambda_{U_j}}^2 \right) }{1-\abs{\lambda_{V_1}}\abs{\lambda_{V_N}}\abs{\lambda_{U_j}}^2} & \cdots & \frac{\left( 1-\abs{\lambda_{V_N}}^2 \right) \left( 1-\abs{\lambda_{U_j}}^2 \right) }{1-\abs{\lambda_{V_N}}^2\abs{\lambda_{U_j}}^2}
		\end{bmatrix} 
	\end{split}
\end{equation}
where $\lambda_{V_k}$ and  $\lambda_{U_l}$ for $k,l \in \left\{ 1,\cdots,N \right\} $ are the eigenvalues of $\mV$ and  $\mU$ respectively.
 
The matrix $\mathbf{\Lambda}_j$ essentially represents the \textit{weight} of each entry in the $N\times N$ matrix $\vb_j^{H}\vb_j$, where $\sum_{j=1}^{N} \vb_j^{H}\vb_j = \mB^{H}\mB = \mE_V^{H}\mW^{T}\mW\mE_V$. Note that $\mW\mE_V$ is the projection of each row of  $\mW$ onto the eigenspace of  $\mV$. This is similar to the case \cite{Ganguli18970}, where the authors obtained the projection of feedforward connection vector onto the eigenspace of recurrent connectivity. However, in this case, these projections does not add up to $1$ unless we constrain $\mW$ to be unitary, in which case  $\mB^{H}\mB = \mI$. However, such a constraint will not be beneficial as in essence, we are restricting the input signal to only be visible to the diagonal elements of the state matrix after transformation by $\mU$ and  $\mV$. This has an attenuation effect on the signal entering the state due to the matrix product, which can affect the information negatively in the long term. In Corollary \ref{C-10}, we show this effect of such constraint by proving a bound on $\mJ_{tot}$ which shows the inability of such constraint to store past information. 
 
Now using $\mathbf{\Lambda}_j$, we can find a form of $\mJ_{tot}$ which is a bit simpler
\begin{equation}
	\begin{split}
		\mJ_{tot} &= \frac{1}{\varepsilon_1\varepsilon_2}\sum_{j=1}^{N} \Tr\left( \mathbf{\Lambda}_j \odot \vb_j^{H}\vb_j \right) 
	\end{split}
\end{equation}
or equivalently, we can represent it as the sum of all sums of weighted diagonal elements of $\vb_j^{H}\vb_j$
 \begin{equation}
	 \label{J-tot-final}
	\begin{split}
	\mJ_{tot} &= \frac{1}{\varepsilon_1\varepsilon_2} \sum_{j=1}^{N} \sum_{k=1}^{N} 	\frac{\left( 1-\abs{\lambda_{V_k}}^2 \right) \left( 1-\abs{\lambda_{U_j}}^2 \right) }{1-\abs{\lambda_{V_k}}^2\abs{\lambda_{U_j}}^2} |b_{jk}|^2
	\end{split}
\end{equation}
where $b_{jk}$ is the corresponding element of matrix $\mB$. This completes the proof.
\end{proof}
\begin{remark}
The form in Eq. \ref{J-tot-final} now can be explored analytically, especially due to the fact that the function of eigenvalues in Eq. \ref{J-tot-final},$\frac{\left( 1-\abs{\lambda_{V_k}}^2 \right) \left( 1-\abs{\lambda_{U_j}}^2 \right) }{1-\abs{\lambda_{V_k}}^2\abs{\lambda_{U_j}}^2}$ can easily seen to be $\le 1$, if $\abs{\lambda_{V_k}}$, $\abs{\lambda_{U_k}}$  $\le 1$, which informally implies that the weights for the elements $|b_{jk}|^2$ are $\le 1$ which can be used to create bounds for $\mJ_{tot}$.
\end{remark}

As visible from Eq. \ref{J-tot-main-final}, the total capacity is dependent on the mixture of eigenvalues of the connectivity matrices, unlike the capacity in normal vector representation networks where the eigenvalues of the recurrent connectivity all add up to 1, thus yielding $\mJ_{{tot}_{rel}} = 1$ relative to the input Fisher information. On the other hand, in the case of our recurrent system in Eq. \ref{dyn-sys}, we have the input Fisher information given through the following lemma.
\begin{lemma} \label{L-7}(Input Fisher Information) The Fisher information received by the state of the recurrent system in Eq. \ref{dyn-sys} with no non-linearity ($f = \text{id}$) at each timestep is given by,
\begin{equation}
	\label{FI-MatGaus}
	\begin{split}
		\mathcal{I}\left( \mW \right) &= \frac{1}{\varepsilon_1\varepsilon_2} \Tr\left( \mW^{T}\mW \right) = \frac{1}{\varepsilon_1\varepsilon_2} \| \mW\|_F^2 
	\end{split}
\end{equation}
where $\| .\|_F$ is the Frobenius norm.
\end{lemma}
\begin{proof}
The Fisher Information for a random variable $\rx$ is given as
\begin{equation}
	\label{FI}
	\begin{split}
		\mathcal{I}\left( \theta \right) &= -\mathbb{E}\left[ \frac{\partial^2}{\partial \theta^2} \log f_{\rx} \left( x;\theta \right)  \right] 
	\end{split}
\end{equation}
where $f_{\rx}\left( x;\theta \right) $ is the density function of $\rx$ under the regularity condition that $\log f_{\rx}(x;\theta)$ is twice differentiable.
\newline\newline
Consider now the timestep $t$, where the input is given to the state via the following,
\begin{equation*}
	\begin{split}
		\mX\left( t \right) = \mW\vs_0  + \rmZ.
	\end{split}
\end{equation*}
Note that $\vs_0 = \vs\left( t-0 \right) $ and $\rmZ \sim \mathcal{MN}_{n\times p}\left( \mathbf{0}, \text{diag}\left( \varepsilon_1 \right) , \text{diag}\left( \varepsilon_2 \right)  \right) $. Hence, $\mX$ will also be a random matrix with mean $\mW\vs_0$ and same covariance matrices,
\begin{equation*}
	\begin{split}
		\rmX\left( t \right)  \sim \mathcal{MN}_{n\times p} \left( \mW\vs_0, \text{diag}\left( \varepsilon_1 \right), \text{diag}\left( \varepsilon_2 \right)   \right).
	\end{split}
\end{equation*}
We can now calculate Fisher information of $\rmX\left( t \right) $ parameterized by $\vs_0$ simply by using Eq. \ref{FI} to get
\begin{equation*}
	\begin{split}
			\mathcal{I}\left( \vs_0 \right) &= -\mathbb{E}\Biggl[ \frac{\partial^2}{\partial \vs_0^2 }\Biggl( -\log\left( \left( 2\pi \right) ^{\frac{1}{2}np} \abs{\text{diag}(\varepsilon_1)}^{\frac{1}{2}p} \abs{\text{diag}(\varepsilon_2)}^{\frac{1}{2}n} \right)\\
			&\;\;\; -\frac{1}{2}\Tr\left( \text{diag}(\varepsilon_2)^{-1} \left( \mX - \mW\vs_0 \right) ^{T}\text{diag}(\varepsilon_1)^{-1}\left( \mX - \mW\vs_0 \right)  \right) \Biggl) \Biggl]\\
			&= -\mathbb{E}\left[ \frac{\partial^2}{\partial\vs_0^2} \left(-\frac{1}{2\varepsilon_1\varepsilon_2}\Tr\left( \mI\left( \mX - \mW\vs_0 \right) ^{T}\mI\left( \mX - \mW\vs_0 \right)  \right) \right)\right]\\
			&= \frac{1}{2\varepsilon_1\varepsilon_2} \mathbb{E}\left[ \frac{\partial^2}{\partial\vs_0^2} \Tr\left( \left( \mX-\mW\vs_0 \right) ^{T}\left( \mX- \mW\vs_0 \right)  \right)  \right] \\
			&= \frac{1}{2\varepsilon_1\varepsilon_2}\mathbb{E}\left[ \Tr\left( \frac{\partial^2}{\partial\vs_0^2} \left( \mX-\mW\vs_0 \right) ^{T}\left( \mX-\mW\vs_0 \right)  \right)  \right]\\
			&= \frac{1}{2\varepsilon_1\varepsilon_2}\mathbb{E}\left[ \Tr\left( 2\mW^{T}\mW \right)  \right]\\
			&= \frac{1}{\varepsilon_1\varepsilon_2} \Tr\left( \mW^{T}\mW \right).
	\end{split}
\end{equation*}
This completes the proof.
\end{proof}

Given the input Fisher information, we now see an interesting case arising from Theorem \ref{T-6}  when $\mU$ and $\mV$ are further assumed to be convergent apart from being normal. In particular, we observe that $g\left( \abs{\lambda_{V_k}}, \abs{\lambda_{U_j}} \right)= \frac{\left( 1-\abs{\lambda_{V_k}}^2 \right) \left( 1-\abs{\lambda_{U_j}}^2 \right) }{1-\abs{\lambda_{V_k}}^2\abs{\lambda_{U_j}}^2} \in (0,1)$ as $0<\abs{\lambda_{V_k}}, \abs{\lambda_{U_j}} < 1$ due to convergent condition and $\frac{1}{\varepsilon_1\varepsilon_2} \sum_{j=1}^{N} \sum_{k=1}^{N} |b_{jk}|^2 = \frac{1}{\varepsilon_1\varepsilon_2}\Tr\left(\mB^{H}\mB  \right) = \frac{1}{\varepsilon_1\varepsilon_2}\Tr\left( \mW^{T}\mW \right) $ due to normal condition. Hence, we arrive at the following corollary of Theorem \ref{T-6}.
\begin{corollary}\label{C-10}
(Capacity of Normal Convergent Matrix Networks) Given the dynamical system \ref{dyn-sys} and that both the connection matrices $\mU$ and $\mV$ are normal as well as convergent, the memory capacity of the system is limited by the following inequality
\begin{equation}
	\label{J-tot-bound-normal}
	\begin{split}
		\mJ_{tot} < \frac{1}{\varepsilon_1\varepsilon_2} \Tr\left( \mW^{T}\mW \right)  
	\end{split}
\end{equation}
which is equivalent to stating that the memory capacity of dynamical system in Equation  \ref{dyn-sys} with respect to instantaneous input Fisher information is
\begin{equation}
\label{capac-rel}
	\begin{split}
		\mJ_{tot_{rel}} = \frac{\mJ_{tot}}{\mathcal{I}(\mW)} < 1
	\end{split}
\end{equation}
where $\mathcal{I}(\mW)$ is the input Fisher information as in Lemma \ref{L-7}.\qed
\end{corollary}

The bound in Eq. \ref{J-tot-bound-normal} implies that the amount of information stored in the state of Eq. \ref{dyn-sys} about all the previous input signals is less than the amount of input information at the current instance. Eq. \ref{capac-rel} further implies that the normal convergent matrix representation in recurrent neural networks cannot efficiently redistribute the total past information with respect to the information just observed. Thus, the memory capacity of such a network would be suboptimal and would not perform well on short/long term recall tasks and also will be likely to fail in generalizing over those respective tasks. A final issue to note about the assumption of convergence is that it is not required in the case of vector representation networks, as the magnitude of all eigenvalues has to be less than 1 due to asymptotic stability criterion \cite{Ganguli18970}.

\subsubsection{Capacity of General Non-Normal Networks}
\label{Cap-NonNormal-Net}

So far, we have seen the inability of normal convergent connectivity in matrix
RNNs in the storage of information about past signals.
However, we have only dealt under the constraint of normal connectivity matrices. It is not yet clear how the performance might change if we relax this constraint (whether $\mJ_{tot}$'s upper bound increases or decreases even further). Relaxing this constraint, however, is not straightforward, as can be seen from the difficulty in the direct analysis of Eqs. \ref{fc} and \ref{J-tot-sum-form}. Under similar challenges, Ganguli et. al \cite{Ganguli18970} defined a broader form of FMM, known as \textit{space-time FMM}.  Here, the temporal signal $\vs_k$ was assumed to be supported also by the spatial dimension introduced by the feedforward connectivity vector for each signal at each time. This formulation depends on the fact that the form of $\mJ_{i,j}$ turns out to be simpler in the case of vector representations as seen in Eq. \ref{fmm-vec-rep}, unlike Eq. \ref{fmm-simplest}. Specifically, the vector feedforward connection  $\vv$ in \ref{fmm-vec-rep} acts on $\mW^{iT}\mC_n^{-1}\mW^{j}$ which can be seen as a matrix storing the information regarding all the elements of $\vv$ (which can be regarded as the spatial dimension) for each time step, which leads to the formulation of $\mJ_{(i,m),(j,n)}^{st} = \left[ \mW^{iT}\mC_n^{-1}\mW^{j} \right]_{(m,n)} $ (keep in mind the different context of $\mW$ for vector dynamics). The introduction of $\mJ^{st}$ makes the analysis of $\mJ\left( i \right) $ for vector representation dynamics for non-normal connectivity much more accessible.

The formulation of an equivalent paradigm in matrix representation dynamics seems to be a non-trivial task; however, especially due to the fact that we can't separate the feedforward connection matrix $\mW$ from the matrix products inside the \textit{trace operator} in Eq. \ref{fmm-simplest}. The presence of $\mW$ in between the products indicates the presence of complicated interaction terms between the feedforward connection and the recurrent connections. This interaction might be a result of using structured matrix representations, which we explain further.

The spatial dimension in matrix representations should correspond to the signal $\vs_i = s\left( n-i \right) $ reaching each neuron in the state matrix $\mX\left( n \right) $. Note that the input signal is a scalar $\vs_k$, thus for vector representations, the $k^{th}$ neuron receives $\vv_k s\left( n-i \right)$ as the signal. Similarly, in matrix representations, the $(k,l)^{th}$ neuron will receive $\mW_{k,l}s\left( n-i \right) $, thus $\mW$ provides structure to input signal to reach the neuron. Without this structural support, the signal can clearly not reach the state neuron. Thus, it seems logical that the capacity of matrix representations would directly depend upon the matrix that stratifies the input signal to each neuron.

We can see this effect if we assume that each recurrent connectivity matrix $\mU$ and  $\mV$ induces their respective spatio-temporal FMM $\mJ^{st}_V$ and $\mJ^{st}_U$, which are similar in structure as in the vector representations as discussed in the beginning. 
Hence  $\mJ^{st}_{V_{(i,j)}}$ and $\mJ^{st}_{U_{(i,j)}}$,  store the Fisher information that $\mX\left( n \right) $ contains about the interference of the input signals which appears $i$ and  $j$ time steps in the past along all their respective spatial dimensions for the respective connection matrices $\mU$ and  $\mV$, as defined below concretely
\begin{equation}
	\label{ST-J-tot}
	\begin{split}
		\mJ^{st}_{V_{(i,j)}} &:= \mV^{j}\mathbf{\Sigma}^{-1}\mV^{iT}\\
		\mJ^{st}_{U_{(i,j)}} &:= \mU^{i}\mathbf{\Psi}^{-1}\mU^{jT}.
	\end{split}
\end{equation}
Using Eq. \ref{ST-J-tot}, we can reform Eq. \ref{fmm-simplest} in a much simpler form which shows the effect of each of these matrices on the $\mJ_{i,j}$. We first write
\begin{equation}
	\label{st-fmm}
	\begin{split}
		\mJ_{i,j} &= \Tr\left( \mW\mJ^{st}_{V_{\left( i,j \right)}} \mW^{T}\mJ^{st}_{U_{\left( i,j \right) }} \right) 
	\end{split}
\end{equation}
and thus, the $\mJ_{tot}$ becomes
 \begin{equation}
	 \label{st-J-tot}
	\begin{split}
		\mJ_{tot} &= \sum_{i=0}^{\infty} \Tr\left( \mW\mJ^{st}_{V_{\left( i,i \right) }}\mW^{T}\mJ^{st}_{U_{\left( i,i \right) }} \right). 
	\end{split}
\end{equation}
Thus, the FMM $\mJ_{i,j}$ depends upon the interaction between each of the spatio-temporal FMM along with the feedforward connection which connects the input signal to the state neurons. Now, since each  spatio-temporal matrix is positive definite and $\varepsilon_2 \sum_{i=0}^{\infty}\Tr\mJ^{st}_{V_{\left( i,i \right) }}  = \varepsilon_1\sum_{i=0}^{\infty}\Tr \mJ^{st}_{U_{\left( i,i \right) }}  = N$, we can retrieve a fundamental bound on the memory capacity $\mJ_{tot}$ of Eq. \ref{dyn-sys} when there are no non-linearity, which we state as the following theorem.
\begin{theorem}\label{T-11}
(Capacity of General Networks)  Given the recurrent system in Eq. \ref{dyn-sys} with no non-linearity ($f(\cdot) = \text{id}(\cdot)$), the memory capacity of the system is fundamentally limited by the following inequality
\begin{equation}
	\label{J-tot-fund-bound}
	\begin{split}
		\mJ_{tot} \le \frac{N^2}{\varepsilon_1\varepsilon_2}\Tr\left( \mW^{T}\mW \right) 
	\end{split}
\end{equation}
which in terms of instantaneous relative capacity is,
\begin{equation}
	\label{J-tot-fund-bound-rel}
	\begin{split}
		\mJ_{{tot}_{rel}} \le N^2.
	\end{split}
\end{equation}
\end{theorem}
\begin{proof}
We have that,
\begin{equation*}
	\begin{split}
		\mJ_{tot} &=  \sum_{i=0}^{\infty} \Tr\left( \mW\mJ_{V_{(i,i)}}^{st} \mW^{T}\mJ^{st}_{U_{(i,i)}}\right)
	\end{split}
\end{equation*}
where $\varepsilon_2\sum_{i=0}^{\infty} \Tr\left( \mJ_{V_{(i,i)}}^{st} \right) = \varepsilon_1\sum_{i=0}^{\infty} \Tr\left( \mJ_{U_{(i,i)}}^{st} \right) = N$. Now to obtain the inequality in Eq. \ref{J-tot-fund-bound}, we proceed as follows,
\begin{equation*}
	\begin{split}
		\mJ_{tot} &=  \sum_{i=0}^{\infty} \Tr\left( \mW \mJ_{V_{(i,i)}}^{st}\left( \mJ_{U_{(i,i)}}^{st}\mW \right) ^{T} \right).
	\end{split}
\end{equation*}
Since the norm induced by Frobenius inner product $\langle \mA, \mB\rangle_F = \Tr\left( \mA^{T}\mB \right) $ will obey Cauchy-Schwarz inequality, i.e. $\abs{\langle \mA, \mB \rangle_F }\le \|\mA\|_F \|\mB\|_F$, we hence get,
\begin{equation*}
	\begin{split}
		\mJ_{tot}&\le \sum_{i=0}^{\infty} \sqrt{\Tr\left( \mW\mJ_{V_{(i,i)}}^{st}\mJ_{V_{(i,i)}}^{st\;T}\mW^{T} \right)\Tr\left( \mW^{T}\mJ_{U_{(i,i)}}^{st}\mJ_{U_{(i,i)}}^{st\;T}\mW \right)  }\\
		&=  \sum_{i=0}^{\infty} \sqrt{\Tr\left(\mW^{T}\mW\mJ_{V_{(i,i)}}^{st}\mJ_{V_{(i,i)}}^{st\;T}\right)\Tr\left( \mW\mW^{T}\mJ_{U_{(i,i)}}^{st}\mJ_{U_{(i,i)}}^{st\;T}\right)  }\\
	\end{split}
\end{equation*}
Now since $\mW^{T}\mW$, $\mJ_{V_{(i,i)}}^{st}\mJ_{V_{(i,i)}}^{st\;T}$ and $\mJ_{U_{(i,i)}}^{st}\mJ_{U_{(i,i)}}^{st\;T}$  are positive semi-definite and $\Tr\left( \mJ_{V_{(i,i)}}^{st} \right) = \Tr\left( \mV^i \mV^{iT} \mathbf{\Sigma}^{-1} \right) \ge 0$ as $\mV^i\mV^{iT}$ and $\mathbf{\Sigma}^{-1}$ are positive semi-definite, hence we can get,
\begin{equation*}
	\begin{split}
		\mJ_{tot} &\le \sum_{i=0}^{\infty} \Tr\left( \mW^{T}\mW \right) \sqrt{\Tr\left(\mJ_{V_{(i,i)}}^{st}\mJ_{V_{(i,i)}}^{st\;T}\right)\Tr\left( \mJ_{U_{(i,i)}}^{st}\mJ_{U_{(i,i)}}^{st\;T}\right)  }\\
		&\le \sum_{i=0}^{\infty} \Tr\left( \mW^{T}\mW \right) \sqrt{\Tr^2\left( \mJ_{V_{(i,i)}}^{st} \right) \Tr^2\left( \mJ_{U_{(i,i)}}^{st} \right) }\\
		&= \Tr\left( \mW^{T}\mW \right)\sum_{i=0}^{\infty} \abs{\Tr\left( \mJ_{V_{(i,i)}}^{st} \right)} \abs{ \Tr\left( \mJ_{U_{(i,i)}}^{st} \right)}\\
		&\le \Tr\left( \mW^{T}\mW \right) \left( \sum_{i=0}^{\infty} \abs{\Tr\left( \mJ_{V_{(i,i)}}^{st} \right)}  \right) \left( \sum_{i=0}^{\infty} \abs{\Tr\left(\mJ_{U_{(i,i)}}^{st}\right)}\right)\\
		&=  \Tr\left( \mW^{T}\mW \right) \frac{N^2}{\varepsilon_1\varepsilon_2}.
	\end{split}
\end{equation*}
This completes the proof.
\end{proof}
Therefore, the memory capacity of a matrix representation of linear recurrent networks is fundamentally bounded by the size of the state matrix. It's interesting to note that besides the apparent flaw in capacity under normal connectivity constraints for matrix representation networks, the capacity in the general case at least seems to be in line with the trend first observed with vector representation recurrent networks, where the $\mJ_{tot_{rel}}\le N$ for $N$ neurons \cite{Ganguli18970}. However, as visible from the case of normal connectivity matrices, the question of whether the system's capacity would be close to this bound is not trivially answered. Clearly, the capacity couldn't reach the bound by optimizing only $\mW$, as  $\mJ_{tot_{rel}}$ doesn't depend on $\mW$. However, one can argue for the case of high $ \|\mU\|_{F}$ and $\|\mV\|_{F}$ which becomes apparent from Eq. \ref{st-J-tot} and the above proof.

In our analysis of the memory capacity of matrix representation networks so far, we assumed that Eq. \ref{dyn-sys} works under linear dynamics; that is, $f(\mX) = \mX$. However, in practice, some type of non-linearity is used for learning efficient and diverse representations. The same can be said for biological neurons as their activations are not unbounded, hence the finite dynamic range. We next show that adding a saturating non-linearity like $\sigmoid(x)$ or $\tanh(x)$ causes the capacity to also be bounded by a multiple of the same range.


\subsection{Effects of finite dynamic range}\label{S-EoFDR}

So far, we've seen the asymptotic effects on the memory capacity of the linear dynamics matrix representation networks. However, it still remains to be answered the effects on the capacity of the system when the dynamics are restricted to some finite dynamic range as in the biological neurons through other saturating non-linearities commonly used in the machine learning community.

\subsubsection{Memory Capacity is upper-bounded by the norm of state}

To see whether such non-linearities increase or decrease the memory capacity, we first assume that the network architecture is such that neural activity of each neuron in $\mX\left( n \right) $ is limited between $-\sqrt{R}$ and $\sqrt{R}$; hence, restricting $\Tr\left( \mX^{T}\mX \right) < N^2R$.

Looking at the average of the state norm of  Eq. \ref{dyn-sys} which is further  shown  in Eq. \ref{avg-norm-state}, we see that $\Tr\left( \mX^{T}\mX \right) < N^2R$ implies that each of its components are also bounded by $N^2R$

\begin{equation*}
	\label{avg-norm-state}
	\begin{split}
		\mathbb{E}\left[ \Tr\left( \mX^{T}\mX \right)  \right] &= \sum_{k=0}^{\infty} \|\mU^{kT}\mW\mV^{k}\|_{F}^2 + \varepsilon_1\varepsilon_2 \sum_{k=0}^{\infty} \Tr\left( \mV^{kT}\mV^{k} \right) \Tr\left( \mU^{kT}\mU^{k} \right) 
	\end{split}
\end{equation*}

which implies that $\sum_{k=0}^{\infty} \|\mU^{kT}\mW\mV^{k}\|_{F}^2  < N^2R$.

On the other hand, since $\mathbf{\Sigma}^{-1}$ and $\mathbf{\Psi}^{-1}$ are positive definite, we see that the capacity \ref{J-tot-mat-rep} of the matrix network is
\begin{equation}
	\begin{split}
	\mJ_{tot} &=  \Tr\left( \sum_{i=0}^{\infty} \mathbf{\Sigma}^{-1}\mV^{iT}\mW^{T}\mU^{i}\mathbf{\Psi}^{-1}\mU^{iT}\mW\mV^{i}\right)\\
	&=  \Tr\left( \mathbf{\Sigma}^{-1} \sum_{i=0}^{\infty} \mV^{iT}\mW^{T}\mU^{i}\mathbf{\Psi}^{-1}\mU^{iT}\mW\mV^{i}\right)\\
	&\le \Tr\left( \mathbf{\Sigma}^{-1} \right) \Tr\left( \mathbf{\Psi}^{-1} \right) \left( \sum_{i=0}^{\infty} \Tr\left( \mV^{iT}\mW^{T}\mU^{i}\mU^{iT}\mW \mV^{i}\right)  \right)\\
	&= \Tr\left( \mathbf{\Sigma}^{-1} \right) \Tr\left( \mathbf{\Psi}^{-1} \right) \left( \sum_{i=0}^{\infty} \|\mU^{iT}\mW\mV^{i}\|_{F}^2 \right)\\
	&\le \Tr\left( \mathbf{\Sigma}^{-1} \right) \Tr\left( \mathbf{\Psi}^{-1} \right) \mathbb{E}\left[ \Tr\left( \mX^{T}\mX \right)  \right]\\
	&< \Tr\left( \mathbf{\Sigma}^{-1} \right) \Tr\left( \mathbf{\Psi}^{-1} \right) N^2R.
\end{split}
\end{equation}

Hence, when we restrict the neuronal dynamic range of activation between certain thresholds, the memory capacity of the network is also limited by that same threshold with appropriate constants. This implies that in the case of saturating non-linearities such as sigmoid or $\tanh$. The memory capacity $\mJ_{tot}$ may decrease whereas, for non-saturating non-linearities such as ReLU and its derivatives, this bound doesn't create a problem from a memory standpoint.

So far, we've seen an analysis of the memory capacity for matrix representation recurrent network as described by the linear dynamics of Eq. \ref{dyn-sys}. We first considered memory capacity under the constraint of normal connectivity matrices, in which we proved the limited capability of such a network. We did this under further convergence assumption by storing information about past signals which turns out to be worse than conventional vector representation of RNNs under the same constraints.

We then further discussed the general case and showed the information stored in the state relative to input information cannot exceed the number of neurons, which thus seems to generalize from the similar results obtained for vector representation recurrent neural network. The question that now naturally arises is whether there are ways in which $J_{tot}$ can be increased. One obvious way of achieving that is through the addition of an external memory resource to the recurrent neural network which on an intuitive level does increase the memory capacity of the network. Even though there has been very in-depth work on the memory of the non-linear vector neural networks, there hasn't been any that extends those ideas to architectures that have external memory available to exploit, let alone the higher-order representations that we deal with in this paper. We thus explore the idea of quantifying the memory capacity of an exceedingly simple memory network with matrix representation through the definitions introduced in previous sections and provide the absolute minimum increase in memory that one would expect from such a memory architecture.

\section{The Effects of External State Memory on State Dynamics}
\label{TEoESMoSD}

We earlier saw the memory capacity shown in Eq. \ref{J-tot-main-final} of 
 the dynamical system  Eq. \ref{dyn-sys}. In this section, we explore one possible path to increase the aforementioned memory capacity.

\subsection{Direct ways to increase the capacity}

We start by first noting the FMM in Theorem \ref{T-2} of matrix networks. In order to increase this notion of memory capacity, we should clearly either \textit{(a)} increase the mean $\mM$ or \textit{(b)} decrease the covariances $\mathbf{\Sigma}$ and $\mathbf{\Psi}$. However, doing either of them is not clear at first sight. To increase the mean $\mM$, we can try adding more terms to it, which can be done easily if we note the state solution in Eq. \ref{state-sol} is obtained via solving the recurrence relation in Eq. \ref{dyn-sys} in the linear case. Hence, if we add any term in the state dynamics (Eq. \ref{dyn-sys}), then we would get the resultant in Eq. \ref{state-sol}. We note
that if we give the state at time step $n$ of RNN with access to past states,
then the state solution would feature information from each of
those past states, thus increasing the terms in the state solution. This answers the question of what we should add to the RNN
state in Equation \ref{dyn-sys} to have a larger mean. Note that such a structure can be named under the well-explored memory networks \cite{weston2014memory}. In memory networks, the state at each time step is also supplemented through a term read from an external memory in the previous time step and the term that is written to memory is usually some transformation of the current state itself \cite{graves2014neural,graves2016hybrid, santoro2016meta}. Here, we assign the external memory to store certain past states, thus assuming no specific structure that determines what should be stored in the memory at the current time step. Note that such memory should have a queue-like behavior. We explore this idea in detail now and derive the corresponding capacity of such an architecture.


\subsection{Addition of State Memory to State Dynamics}
\label{AoSMtSD}

In this section, we modify Eq. \ref{dyn-sys} by adding a generic finite queue memory that stores the past state representations for the current state as an additional input. The motivation behind this is to access past representations for the current state in order to simplify the task of encoding each past input signal for that state. The main goal now is to find the best representation only for a small previous neighborhood of current input signal instead of all past signals, while encoding minimal to no information for those past signals as they are readily available through this type of memory. We further show that even with a bounded memory size of just one slot, the memory capacity becomes an infinite matrix series sum which is at the very least upper bounded by four times the capacity of the matrix representation networks without memory access at each time step as given in Eq. \ref{fc}. Hence, we mathematically prove the increase in capacity such a memory structure will have for matrix representation in RNNs.

\subsubsection{The change in dynamics}
Consider the following recurrent system
\begin{equation}
	\label{mem-dyn-sys}
	\begin{split}
		\mX\left( n \right) &= \mU^{T} \mX\left( n-1 \right) \mV + \mW s\left( n \right) + \mQ_{\text{READ}}\left( n \right) + \rmZ\left( n \right)  
	\end{split}
\end{equation}
where $\mU$,  $\mV$,  $\mW$ and  $\rmZ$ have the same meaning as in Eq. \ref{dyn-sys}. The  $\mQ_{\text{READ}}\left( n \right) $ represents the matrix read from memory at time $n$, which in general can be a function on the set of all the past elements of memory $\left\{ \mQ\left( t \right): 0\le t<n-1 \right\} $. In practice, most of the memory-based neural architectures only consist of a finite memory span while the read operation mapping from the memory $\mQ$ to an extracted read element happens through a key-value retrieval mechanism. In most cases, this map is linear where a scalar strength is attributed to each \textit{slot} of memory based on a key usually generated through an RNN \cite{graves2014neural,santoro2016meta}.

In the same motivation, one can define a queue-like memory $\mQ_n$ of size $p$, where $\mQ_n[i] \in \mathbb{R}^{N\times N}$ for all $i \in \left\{ 1,\cdots,p \right\}$ stores the state $\mX\left( n-1-i \right)$; i.e. $\mQ_n\left[ i \right] = \mX(n-1-i)$. The update to memory happens as an \verb|enqueue()| operation which adds $\mX\left( n-1 \right)$ as a new slot on the front and at the same time the operation \verb|dequeue()| removes the slot $\mQ_n[p]$ containing the state $\mX\left( n-1-p \right) $. To read from $\mQ_n$, we describe a sequence of scalars $\left\{ \alpha_k \right\}$ for $k \in {1,\cdots,p}$ which describes the strength of each of the memory locations in the current time step $n$, such that $\sum_{k=1}^{p} \alpha_k = 1$. The sequence $\{\alpha_k\}$ can be functions of the previous state which can be dependent on the input signal or just be given as a constant (perhaps, they may give equal strength to all slots for time   $n$). Given this construct, we can now define the $\mQ_\text{READ}\left( n \right) $ formally as
\begin{equation}
	\label{M-READ}
	\begin{split}
		\mQ_{\text{READ}}\left( n \right) &:= \sum_{t=1}^{p} \alpha_t \mQ_n[t] = \sum_{t=1}^{p} \alpha_t \mX\left( n-1-t \right).
	\end{split}
\end{equation}
We now see how the memory capacity of the system in Eq. \ref{mem-dyn-sys} is altered in comparison to the system in Eq. \ref{dyn-sys} which will be further discussed in the next section. 

\subsubsection{Calculating the altered FMC in a simple case}

Given the memory structure in place, we now state  
the FMC in Eq. \ref{mem-dyn-sys} and then prove it in the discussion that follows.
\begin{theorem}\label{T-12}
(FMC of Matrix Memory Network) The Fisher Memory Curve of the recurrent system \ref{mem-dyn-sys} with $p=1$ is
\begin{equation*} 
	\begin{split}
		\mJ\left( k \right)^{\prime} = \Tr \Biggl( \mathbf{\Sigma}^{-1}_{\text{MF}}&\mathbf{\Sigma}^{-1}_{\text{STATE}}\left( \partial_{\vs_k}\mM_{\text{MEM}}^{T}  \right) \mathbf{\Psi}^{-1}_{\text{MF}} \mathbf{\Psi}^{-1}_{\text{STATE}} \left( \partial_{\vs_k}\mM \right)  \\
		&+\;\mathbf{\Sigma}^{-1}_{\text{MF}}\mathbf{\Sigma}^{-1}_{\text{STATE}}\left( \partial_{\vs_k}\mM^{T}  \right) \mathbf{\Psi}^{-1}_{\text{MF}} \mathbf{\Psi}^{-1}_{\text{STATE}} \left( \partial_{\vs_k}\mM_{\text{STATE}} \right) \Biggl)
		\end{split}
\end{equation*}
\textit{where,}
\begin{equation*}
	\label{mem-frac}
	\begin{split}
		\mathbf{\Sigma}^{-1}_{\text{MF}} &:= \mI - \mathbf{\Sigma}^{-1}_{\text{STATE}} \left( \mathbf{\Sigma}^{-1}_{\text{MEM}} + \mathbf{\Sigma}^{-1}_{\text{STATE}} \right)^{-1}\\
		\mathbf{\Psi}^{-1}_{\text{MF}} &:= \mI - \mathbf{\Psi}^{-1}_{\text{STATE}} \left( \mathbf{\Psi}^{-1}_{\text{MEM}} + \mathbf{\Psi}^{-1}_{\text{STATE}} \right)^{-1}
	\end{split}
\end{equation*}
\textit{where} $\mathbf{\Sigma}^{-1}_{\text{STATE}}$, $\mathbf{\Sigma}^{-1}_{\text{MEM}}$, $\mathbf{\Psi}^{-1}_{\text{STATE}}$ \textit{and} $\mathbf{\Psi}^{-1}_{\text{MEM}}$ are given by Eq. \ref{mean-cov-mem-dyn-sys} below.
\end{theorem}

\begin{proof}
The addition of memory to Eq. \ref{dyn-sys} comes with obvious benefits as discussed earlier. However, proving these effects mathematically might give us a clue about the limits of such effects. To derive the FMC for Eq. \ref{mem-dyn-sys}, we first note that an attempt at finding a general solution of Eq. \ref{mem-dyn-sys} yields the following recurrence relation
\begin{equation}
	\label{mem-dyn-sys-sol}
	\begin{split}
		\mX\left( n \right) &= \sum_{k=0}^{\infty} \mU^{kT}\mW\vs_k\mV^{k} + \sum_{k=0}^{\infty} \mU^{kT}\rmZ\left( n-k \right) \mV^{k}\\ &\;\;\;+ \sum_{k=0}^{\infty} \mU^{kT} \underbrace{\left( \sum_{t=1}^{p} \alpha_t \mX\left( n-k-1-t \right)  \right) }_{\mQ_{\text{READ}}\left( n-k \right) }\mV^{k}
	\end{split}
\end{equation}

where we see an extra term as compared to Eq. \ref{state-sol}. It's not difficult to see how the inclusion of the sum of $\mQ_{\text{READ}}\left( n-k \right) $ for infinitely many $k$ makes the state equation difficult to analyze. We need to recursively substitute $\mX\left( n-k-\cdots \right) $ infinitely many times, each one being an infinite sum itself. However, we can simplify the analysis by making a very reasonable and widely used assumption that more memory (slots) will lead to much broader access to the past states and thus, will lead to better performance. Given this simple assumption, it only becomes necessary for us to show the effectiveness of $p=1$ or just $1$ slot of memory which can, due to our assumption, act as a lower bound on the memory capacity thus derived. Hence, we now only analyze the simple but effective case of $p=1$. Note that in this case, $\{\alpha_k\}=\alpha_1$ and $\alpha_1 = 1$.
 \newline\newline
 With $p=1$, we can re-write \ref{mem-dyn-sys-sol} as follows
\begin{equation}
	\label{mem-gen-sol-1}
	\begin{split}
		\mX\left( n \right) &= \sum_{k=0}^{\infty} \mU^{kT}\mW\vs_k\mV^{k} + \sum_{k=0}^{\infty} \mU^{kT}\rmZ\left( n-k \right) \mV^{k}\\ &\;\;\; + \sum_{i_1 = 2}^{\infty} \mU^{(i_1-2)T} \mX\left( n-i_1 \right)\mV^{i_1-2}.
	\end{split}
\end{equation}

For ease of representation, let us define $\mA=\sum_{k=0}^{\infty}\mU^{kT}\mW\vs_k\mV^{k}$ and $\mB^{n}=\sum_{k=0}^{\infty}\mU^{kT}\rmZ\left( n-k \right) \mV^{k}$. Therefore, we can now substitute the value of $\mX\left( n-i_1 \right) $ in Eq. \ref{mem-gen-sol-1}  to get the following form
\begin{equation}
	\label{mem-gen-sol-2}
	\begin{split}
		\mX\left( n \right) = \mA &+ \sum_{i_1 = 2}^{\infty} \mU^{(i_1-2)T}\mA\mV^{i_1-2}\\ &+ \sum_{i_1 = 2}^{\infty} \sum_{i_1=2}^{\infty} \mU^{(i_1 + i_2 - 4)T} \mX\left( n-i_1-i_2 \right) \mV^{i_1+i_2-4} \\
		&+ \sum_{i_1=2}^{\infty} \mU^{(i_1-2)T}\mB^{n-i_1}\mV^{i_1-2} +\mB^{n}.
	\end{split}
\end{equation}

It's now trivial to see the above's extension to infinite sums of combinations of $\mA$ and  $\mB^{(n-i_1-i_2-\ldots)}$. Before that, let us define the following
\begin{equation}
	\label{SA-SB}
	\begin{split}
		S_\mA\left( m \right) &= \sum_{i_m = 2}^{\infty} \cdots \sum_{i_1=2}^{\infty} \mU^{\left( \sum_{j=1}^{m} i_j - 2m \right)T } \mA \mV^{\left( \sum_{j=1}^{m} i_j - 2m \right)}\\
		S_\mB\left( m \right) &= \sum_{i_m = 2}^{\infty} \cdots \sum_{i_1=2}^{\infty} \mU^{\left( \sum_{j=1}^{m} i_j - 2m \right)T } \mB^{\left( n - \sum_{j=1}^{m} i_j \right) } \mV^{\left( \sum_{j=1}^{m} i_j - 2m \right)}
	\end{split}
\end{equation}

where $S_\mA\left( 0 \right) = \mA $ and $S_\mB\left( 0 \right) = \mB^{n}$. Given Eq.  \ref{SA-SB}, we can now present the general solution of $\mX\left( n \right) $ as the following
\begin{equation*}
	\label{mem-gen-sol-final}
	\begin{split}
		\mX\left( n \right) &= \sum_{m=0}^{\infty} \left[  S_\mA\left( m \right) + S_\mB\left( m \right) \right].
	\end{split}
\end{equation*}

Using the same arguments as in Section \ref{The-FMC}, the mean and covariance matrices for $p\left( \mX\left( n \right) \vert \vs \right)$ can be easily seen to be (note that both rows and columns of $\rmZ\left( n-k \right)$ are independent)
\begin{equation}
	\label{mean-cov-mem-dyn-sys}
	\begin{split}
		\mM\left( \vs \right)  &= \underbrace{\sum_{m=1}^{\infty} \sum_{i_m=2}^{\infty} \cdots \sum_{i_1=2}^{\infty} \sum_{k=0}^{\infty} \mU^{\left( \sum_{j=1}^{m} i_j -2m + k \right)T } \mW \mV^{\left( \sum_{j=1}^{m} i_j - 2m + k \right) } \vs_k}_{\mM_{\text{MEM}}}\\ &\;\;\;+ \underbrace{\sum_{k=0}^{\infty} \mU^{kT}\mW\mV^{k}\vs_k}_{\mM_{\text{STATE}}}\\ 
		\mathbf{\Psi} &= \underbrace{\varepsilon_1 \sum_{m=1}^{\infty} \sum_{i_m=2}^{\infty} \cdots \sum_{i_1=2}^{\infty} \sum_{k=0}^{\infty} \mU^{\left( \sum_{j=1}^{m} i_j -2m + k \right)T } \mU^{\left( \sum_{j=1}^{m} i_j -2m + k \right) }}_{\Psi_{\text{MEM}}}\\ &\;\;\; + \underbrace{\varepsilon_1\sum_{k=0}^{\infty} \mU^{kT}\mU^{k}}_{\Psi_{\text{STATE}}} \\
		\mathbf{\Sigma} &= \underbrace{\varepsilon_2 \sum_{m=1}^{\infty} \sum_{i_m=2}^{\infty} \cdots \sum_{i_1=2}^{\infty} \sum_{k=0}^{\infty} \mV^{\left( \sum_{j=1}^{m} i_j - 2m + k\right)T} \mV^{\left( \sum_{j=1}^{m} i_j-2m+k \right) }}_{\Sigma_{\text{MEM}}} \\ &\;\;\; + \underbrace{\varepsilon_2\sum_{k=0}^{\infty} \mV^{kT}\mV^{k}}_{\Sigma_{\text{STATE}}}.
	\end{split}
\end{equation}

We refer back to Eq. \ref{mean-cov-mem-dyn-sys} and note that the addition of just one slot of state memory to the state changes the memory dynamics hugely. We can further classify the mean $\mM$ and covariance matrices $\mathbf{\Psi}$, $\mathbf{\Sigma}$ into two components identified from the origin of the contribution, whether the contribution comes from the addition of memory or the memory innately to state itself which we have discussed earlier. Such a classification can help us to understand the scale of contribution to the memory capacity of the dynamics (Eq. \ref{mem-dyn-sys}) by each of the two classes. Note that the case when $m=0$ just corresponds to the case when there's no state memory attached to the dynamics.

Deriving the FMC is straightforward now, considering Eq. \ref{mean-cov-mem-dyn-sys}, as stated below
\begin{equation}
	\label{fmc-mem}
	\begin{split}
		\mJ_{k,k}^{\prime} =\mJ\left( k \right)^{\prime} = \Tr\left( \mathbf{\Sigma}^{-1}\frac{\partial \mM\left( \vs \right)^{T} }{\partial\vs_k}  \mathbf{\Psi}^{-1} \frac{\partial \mM \left( \vs \right) }{\partial \vs_k}\right)  
	\end{split}
\end{equation}

where
\begin{equation}
	\label{mean-deriv}
	\begin{split}
		\partial_{\vs_k} \mM &= \frac{\partial \mM\left( \vs \right) }{\partial \vs_k}\\& = \left(\underbrace{\sum_{m=1}^{\infty}  \sum_{i_m=2}^{\infty} \cdots \sum_{i_1=2}^{\infty} \mU^{\left( \sum_{j=1}^{m} i_j -2m +k \right) T} \mW \mV^{\left( \sum_{j=1}^{m} i_j -2m + k \right) }}_{\partial_{\vs_k} \mM_{\text{MEM}}} + \mU^{kT}\mW\mV^{k}\right)
	\end{split}
\end{equation}

Looking at Eq. \ref{fmc-mem} from the context of Eq. \ref{mean-deriv}, we can see informally that due to the \textit{heavy} contribution from the memory towards all the measures in Eq.  \ref{mean-cov-mem-dyn-sys}, the  $\mJ_{k,k}^{\prime}$ seems to be much general and larger than $\mJ_{k,k}$ encountered earlier in the case without memory structure $\mQ$ (Eq. \ref{fc}). More formally, using Eq. \ref{mean-cov-mem-dyn-sys}, we get the following long form of FMC $\mJ\left( k \right) ^{\prime}$

\begin{equation}
	\label{fmc-long}
	\begin{split}
		\mJ\left( k \right) ^{\prime} = \Tr\Biggl( &  \mathbf{\Sigma}^{-1} \left( \partial_{\vs_k} \mM_{\text{MEM}}^{T}\right) \mathbf{\Psi}^{-1}\left( \partial_{\vs_k} \mM_{\text{MEM}}  \right)\\ &+ \mathbf{\Sigma}^{-1}\mV^{kT}\mW^{T}\mU^{k}\mathbf{\Psi}^{-1}\left( \partial_{\vs_k} \mM_{\text{MEM}} \right) \\
		&+ \mathbf{\Sigma}^{-1} \left( \partial_{\vs_k} \mM_{\text{MEM}}^{T}\right) \mathbf{\Psi}^{-1} \mU^{kT}\mW\mV^{k}\\& + \underbrace{\mathbf{\Sigma}^{-1}\mV^{kT}\mW^{T}\mU^{k}\mathbf{\Psi}^{-1}\mU^{kT}\mW\mV^{k}}_{\mJ_1^{\prime} } \Biggl) 
	\end{split}
\end{equation}

where $\mJ_1^{\prime}$ can be simplified further by using \textit{Woodbury matrix identity}\footnote{Consider conformable matrices $\mA$, $\mU$,  $\mC$ and $\mV$, then
\begin{equation*}
	\begin{split}
		\left( \mA+\mU\mC\mV \right)^{-1} &= \mA^{-1} - \mA^{-1}\mU\left( \mC^{-1} + \mV\mA^{-1}\mU \right)^{-1}\mV \mA^{-1}. 
	\end{split}
\end{equation*}} \cite{woodbury1950inverting} on $\mathbf{\Sigma}^{-1}$ and $\mathbf{\Psi}^{-1}$ which yields the following form of covariance matrices

\begin{equation}
	\label{cov-woodbury}
	\begin{split}
		\mathbf{\Sigma}^{-1} &= \mathbf{\Sigma}_{\text{STATE}}^{-1} -\underbrace{\mathbf{\Sigma}_{\text{STATE}}^{-1}\left( \mathbf{\Sigma}_{\text{MEM}}^{-1}  +\mathbf{\Sigma}_{\text{STATE}}^{-1}   \right)^{-1} \mathbf{\Sigma}_{\text{STATE}}^{-1}}_{\Sigma^{-1}_{\text{COMB}}}\\
\mathbf{\Psi}^{-1} &= \mathbf{\Psi}_{\text{STATE}}^{-1} -\underbrace{\mathbf{\Psi}_{\text{STATE}}^{-1}\left( \mathbf{\Psi}_{\text{MEM}}^{-1}  +\mathbf{\Psi}_{\text{STATE}}^{-1}   \right)^{-1} \mathbf{\Psi}_{\text{STATE}}^{-1}}_{\Psi^{-1}_{\text{COMB}}}
	\end{split}
\end{equation}

where $\mathbf{\Sigma}^{-1}_{\text{COMB}}$ describes the part of the inverse apart from $\mathbf{\Sigma}^{-1}_{\text{STATE}}$ where the covariance contributed by addition of memory and the state itself are combined, similarly for $\mathbf{\Psi}^{-1}$. We can now use Eq.  \ref{cov-woodbury} to show that the FMC $\mJ\left( k \right) $ discussed earlier for matrix representation networks (without the above discussed queue-like state memory $\mQ$) is indeed a small part of $\mJ\left( k \right) ^{\prime}$, because $\mJ^{\prime}_1$ now decomposes into the following terms
\begin{equation}
	\begin{split}
		\mJ^{\prime}_1 &=  \underbrace{\mathbf{\Sigma}_{\text{STATE}}^{-1} \mV^{kT}\mW^{T}\mU^{k}\mathbf{\Psi}_{\text{STATE}}^{-1}\mU^{kT}\mW\mV^{k}}_{\mJ\left( k \right) \text{w/o $\Tr$ operator} }\\ &\;\;\; - \mathbf{\Sigma}_{\text{COMB}}^{-1} \mV^{kT}\mW^{T}\mU^{k}\mathbf{\Psi}_{\text{STATE}}^{-1}\mU^{kT}\mW\mV^{k}\\ &\;\;\;- \mathbf{\Sigma}_{\text{STATE}}^{-1} \mV^{kT}\mW^{T}\mU^{k}\mathbf{\Psi}_{\text{COMB}}^{-1}\mU^{kT}\mW\mV^{k}\\ &\;\;\; + \mathbf{\Sigma}_{\text{COMB}}^{-1} \mV^{kT}\mW^{T}\mU^{k}\mathbf{\Psi}_{\text{COMB}}^{-1}\mU^{kT}\mW\mV^{k}    
	\end{split}
\end{equation} 

where we can clearly see the presence of FMC $\mJ\left( k \right) $ of matrix networks without state memory $\mQ$ in the FMC for those who consist of $\mQ$.  In fact, there are three more independent $\mJ\left( k \right) $ which can be seen in Eq. \ref{fmc-long} in each of the terms excluding $\mJ_1^{\prime}$.

Hence, the addition of more terms along with $\mJ\left( k \right) $ to $\mJ\left( k \right) ^{\prime}$ is evident from the addition of the queue-like memory $\mQ$. However, we also note the difficulty of proving more general bounds for the capacity of Eq. \ref{mem-dyn-sys} unlike Eq. \ref{dyn-sys}, which occurs mainly due to inclusion of complicated infinite sums over $m$, where each sum is itself a distorted version of $\mJ\left( k \right) $, as visible from \ref{fmc-long} which do not lend itself to same analysis techniques derived in this paper so far and in Ganhuli et. al \cite{Ganguli18970}; hence, more work might be needed in this direction.

However, we can still generalize Eq. \ref{fmc-long} by using Woodbury  identity again on all four terms inside $\Tr$ operator instead of just $\mJ^{\prime}_1$ which yields us the following
\begin{equation}
	\label{fmc-mem-general}
	\begin{split}
		\mJ\left( k \right)^{\prime} = \Tr \Biggl(& \mathbf{\Sigma}^{-1}_{\text{MF}}\mathbf{\Sigma}^{-1}_{\text{STATE}}\left( \partial_{\vs_k}\mM_{\text{MEM}}^{T}  \right) \mathbf{\Psi}^{-1}_{\text{MF}} \mathbf{\Psi}^{-1}_{\text{STATE}} \left( \partial_{\vs_k}\mM \right)  \\
		&+\mathbf{\Sigma}^{-1}_{\text{MF}}\mathbf{\Sigma}^{-1}_{\text{STATE}}\left( \partial_{\vs_k}\mM^{T}  \right) \mathbf{\Psi}^{-1}_{\text{MF}} \mathbf{\Psi}^{-1}_{\text{STATE}} \left( \partial_{\vs_k}\mM_{\text{STATE}} \right) \Biggl)
		\end{split}
\end{equation}
where, 
\begin{equation}
	\begin{split}
		\mathbf{\Sigma}^{-1}_{\text{MF}} &= \mI - \mathbf{\Sigma}^{-1}_{\text{STATE}} \left( \mathbf{\Sigma}^{-1}_{\text{MEM}} + \mathbf{\Sigma}^{-1}_{\text{STATE}} \right)^{-1}\\
		\mathbf{\Psi}^{-1}_{\text{MF}} &= \mI - \mathbf{\Psi}^{-1}_{\text{STATE}} \left( \mathbf{\Psi}^{-1}_{\text{MEM}} + \mathbf{\Psi}^{-1}_{\text{STATE}} \right)^{-1}.
	\end{split}
\end{equation}
This completes the proof.
\end{proof}
Here, $\mathbf{\Sigma}^{-1}_{\text{MF}}$ can be understood as the fraction of precision provided by memory $\mQ$ via $\mathbf{\Sigma}^{-1}_{\text{MEM}}$, similarly for $\mathbf{\Psi}^{-1}_{\text{MF}}$. Note that $\mathbf{\Sigma}^{-1}_{\text{STATE}}$ and $\mathbf{\Psi}^{-1}_{\text{STATE}}$ are the same set of precision matrices as in the case when there was no memory $\mQ$ given by Theorem \ref{T-2}. Due to this, one can say that addition of queue-like state memory $\mQ$ marginalizes the precision to \textit{effective} precision matrices $\mathbf{\Sigma}^{-1}_{\text{MF}}\mathbf{\Sigma}^{-1}_{\text{STATE}}$ and $\mathbf{\Psi}^{-1}_{\text{MF}}\mathbf{\Psi}^{-1}_{\text{STATE}}$ based on the respective precision contributed due to inclusion of $\mQ$, i.e.  $\mathbf{\Sigma}^{-1}_{\text{MEM}}$ and $\mathbf{\Psi}^{-1}_{\text{MEM}}$.

Given the point of view of the FMC $\mJ\left( k \right)^{\prime}$ with respect to mean $\mM$, the $\mJ\left( k \right) ^{\prime}$ is divided among the contribution from the mixture of the derivative of the complete mean $\mM$ and the contribution to mean via memory $\mM_{\text{MEM}}$ and from the derivative of $\mM$ again with mean contributed via state  $\mM_{\text{STATE}}$. The presence of derivatives of different parts of mean $\mM$ in each of the matrices inside the trace operator in Eq. \ref{fmc-mem-general} unlike Corollary \ref{C-4} in which the derivative of the same mean was taken both times. This implies a stark difference in the underlying structure of $\mJ\left( k \right) ^{\prime}$ compared to $\mJ\left( k \right) $ where the same derivative is taken in the only term present inside the trace operator.  

We note that the \textit{space-time} analogy introduced previously in Section \ref{Cap-NonNormal-Net} would not be feasible with Eq. \ref{fmc-mem-general}   
and hence combined with the above discussion on marginalization of covariance, in this context, deriving a general bound is not trivial. Due to our first assumption that the capacity would increase with more amount of slots $p$, we thus conclude that adding the queue-like state memory $\mQ$ increases the \correction{memory} capacity of 
matrix RNNs.

In summary, we introduced the queue-like state memory $\mQ$ to the state dynamics as defined by Eq. \ref{mem-dyn-sys}. We then derived the FMC for such a memory augmented dynamics and showed its complex structure, showing that the worst-case upper bound of the capacity with just one slot of state memory is 4 times higher than that of the matrix networks without memory. Finally, we gave more insights into the structure of the FMC by deriving even a general form of FMC (Eq. \ref{fmc-mem-general}) and posing some insights about its basic structure.

\section{Matrix Representation in Neural Memory of Recurrent Neural Networks}
\label{MRS-NM}

In the previous section, we revealed how the addition of state memory to a matrix RNNs will affect the memory capacity of the system. In particular, we derived the FMC $\mJ\left( i \right) ^{\prime}$ to evaluate the memory capacity $\mJ_{tot}^{\prime}$ and argued that it is larger than the $\mJ_{tot}$ as derived earlier for matrix representation networks without the memory structure $\mQ_n$ (Theorem \ref{T-11}).

We need to note that a higher memory capacity doesn't necessarily imply that a neural network architecture can efficiently transfer stored knowledge in its hidden representations to achieve the desired output. The memory structure $\mQ_n$ introduced earlier  in Section \ref{AoSMtSD}  
only serves the purpose of increasing the memory capacity, not the efficiency of representations in encoding past inputs clearly \cite{bengio2013representation}.  

It thus can be argued that the dynamical system in Eq. \ref{mem-dyn-sys} is not suited for real-life tasks where one immediate problem often faced is the similarity of input sequences fed across a considerably long time span. One would expect the memory structure $\mQ_n$ to handle such cases and in fact, use the previously seen representation of the same input in the current time step to generate the new representation for the current time step. Hence, an obvious flaw with this structure is the lack of coherence of the hidden representations.  Now for a similar input, we have two different representations, one when the first input arrived long ago in the past and the new \textit{updated} representation which is a result of current similar input and that past state. However, this will only be the case if the past representation is not $\verb|deque()|$ out of the memory $\mQ_n$. That is, the time span between the similar signals is not more than the memory size  $p$ of  $\mQ_n$ and even if the time span between the inputs is less than the memory size, the sequence $\left\{ \alpha_n \right\} $ should be such that all weight is given to the slot, where the past representation is currently stored which can only be the case if $\left\{ \alpha_n \right\} $ is learned to look at the correct \textit{address} in the memory $\mQ_n$. It can also be seen that this problem is only exaggerated if there are multiple similar inputs in the input sequence whose periodicity is greater than the memory size $p$ of $\mQ_n$ which can be the case in recall and copy tasks.

There can be multiple ways to solve the above problem. The memory can act as a placeholder for storing similar or correlated input representation in one slot only, thus all the information about similar inputs is available in one place and can be decoded accordingly, perhaps like an encoder-decoder structure \cite{cho2014properties}. Thus, instead of storing the current state in its entirety, one can instead use it to determine the closest representation already available in the memory and the updates it may need to fit the representation of a similar group of inputs. We can use weights as a unique address of each slot for finding the location of the closest representation in memory. Hence, the weights should somehow be able to look at each slot and determine which one is closest to the current representation.

We note that this problem of finding techniques for efficient storage and retrieval of memory elements is not new in the machine learning research community and there have been multiple proposals for achieving efficient storage and retrieval for different domains \cite{santoro2016meta,gulchere2017memory, le2019neural}. One of the simplest of these proposals is the Neural Turing Machine (NTM) and the  
 memory addressing mechanism it proposes. We thus propose to use the same addressing mechanisms for the memory which now stores matrix representations \cite{gao2016matrix} as opposed to vector representations as originally proposed in NTM. This leads to our proposed  Matrix Neural Turing Machine (MNTM) which takes in and stores matrix-sized sequences into a bounded memory. We now formally introduce MNTM and later argue that even though the modifications are in the overall memory structure as compared to the simple $\mQ_n$ introduced earlier for theoretic evaluations of the effect of memory, the capacity doesn't decrease in comparison while making matrix representations and storage feasible.


\subsection{Matrix Neural Turing Machine}

Storing the  feature vector  
in a differentiable memory can be traced back to Das et. al  \cite{das1992learning}.  However, storing structured representation has been a recent endeavor \cite{pham2018graph, khasahmadi2020memory}. We now introduce 
 a neural memory that stores matrix representations in which addressing is done in the same way as in NTM.

\begin{figure*}[htpb]
	\centering
	\includegraphics[width=1.8\columnwidth]{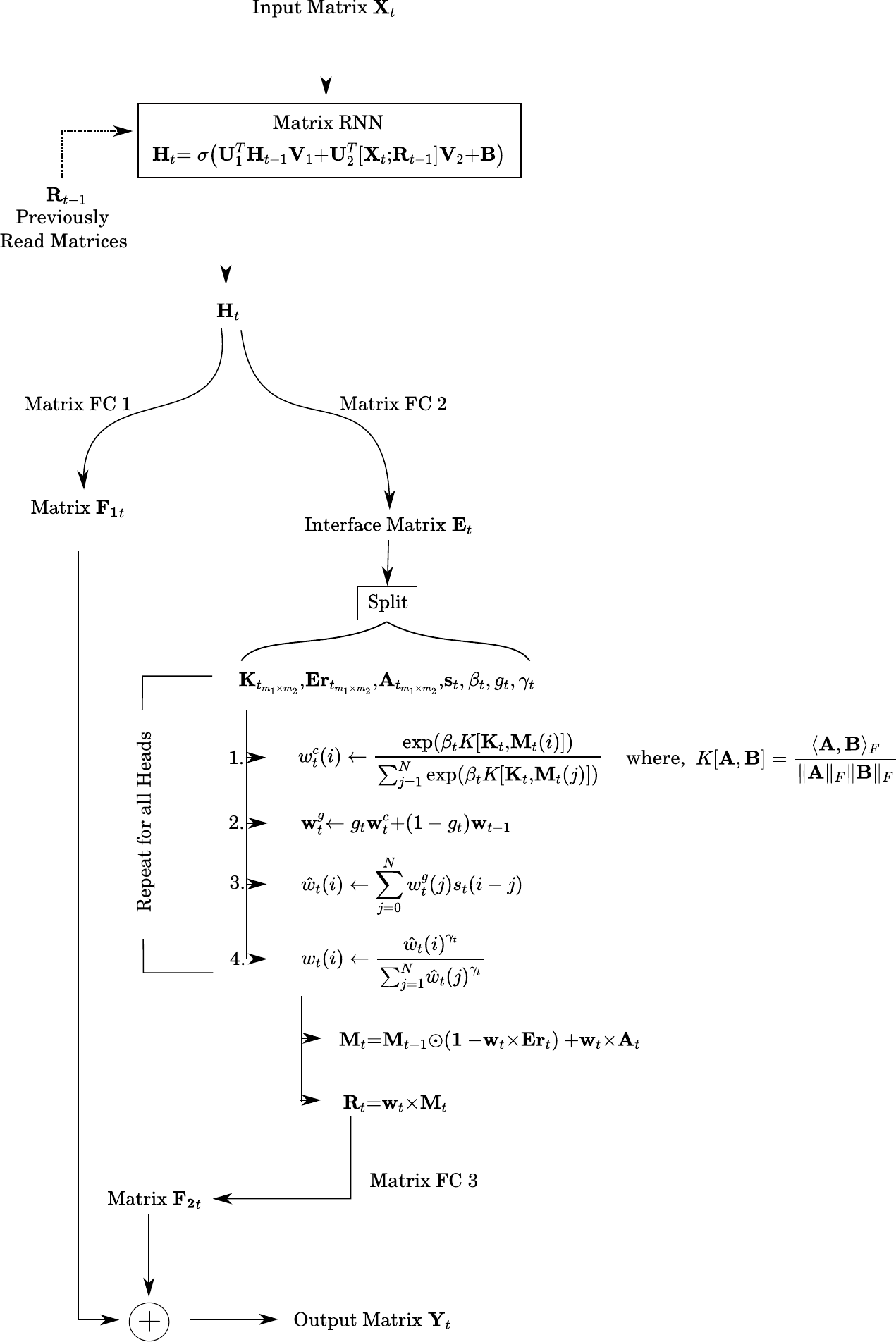}
	\caption{Complete structure of MatNTM. As visible, the whole architecture can be thought of as a matrix analogue of the usual NTMs with a matrix RNN controller network. Matrix FC $\implies$ Matrix Fully Connected layer.}
	\label{fig:matntm}
\end{figure*}

Consider a sequence of input matrices $\left\{ \mX_n \right\}$, and at time $t$, the matrix  $\mX_t$ is given as an input to the matrix RNN that is described in Figure \ref{fig:matntm}. The matrix RNN also receives the matrices read from memory at time $t-1$,  i.e. $\mR_{t-1}$. Since we use only one Read and Write Head, we only receive one matrix read from past memory. The state $\mH_t$ thus generated by matrix RNN is now used to further generate more elements which determine what to add to and delete from the closest representation to the current input available in the memory so that it can be modified to also fit the current representation. The comparison is done through a key $\mK_t$ generated to compare the current input's representation to the representations available in memory.   Hence, we generate the read matrix $\mR_t$  using both addressing by content and addressing by location methods to generate the weights $\vw_t$ for each memory slot.  

The addressing by content (Step 1. in Figure \ref{fig:matntm}) allows us to find the index of the memory closest to key $\mK_t$, where the similarity measure $K\left[ .,. \right] $  is the cosine similarity. The addressing by location (Steps 2-4 in Figure \ref{fig:matntm}) would allow for iterative shifts of the weights, which is an important feature in addition to the content addressing. This will allow us to have an alternative 
 mechanism to the content addressing mechanism in the case when there is no clear representation in the memory closest to the key. Hence, we can avoid adding noise to any of the memory slots by shifting the previous attention or the current closest slot index determined by the content addressing mechanism by a pre-defined number of steps (usually one in either direction).

In reference to the theory explained previously in Sections \ref{TEoESMoSD} and \ref{MRS-NM}, we ought to only care about the retrieval and addition mechanism for memory $\mM_t$. The weights $\vw$ (equivalent to $\left\{ \alpha_n \right\} $) are retrieved via the (learned) state itself which doesn't matter much as discussed earlier in this section. However, the representations stored in memory are the ones with striking differences with the theoretic $\mQ_n$, where we only stored the past states altogether. In order to see the difference more concretely, we can isolate the \textit{erase} $\mE\vr_t$ and \textit{addition} $\mA_t$ matrices for a particular time step $t$ as the following
 \begin{equation*}
	\begin{split}
		\mE\vr_t &=  \mU^{T}_E\mH_t\mV_E\\
		\mA_t &= \mU^{T}_A \mH_t \mV_A
	\end{split}
\end{equation*}
where $\mU_E$,  $\mV_E$ and  $\mU_A$,  $\mV_A$ are the corresponding connection matrices for $\mE\vr_t$ and $\mA_t$ (Figure \ref{fig:matntm}). Note, we only exclude the bias term for comparison of the analysis with  $\mQ_n$. Therefore, in order to analyze the memory retrieval mechanism of MNTM more clearly, and ensure comparability with  $\mQ_n$, we make an important assumption that the number of slots in memory  $\mM_t$ is one for all $t$. Given this assumption in hand, we can thus see that $\mR_t = \mM_t$, and thus we get the following recurrence relation (if we consider that initial memory is all zero)
 \begin{equation}
	 \label{NTM-access}
	\begin{split}
		\mR_t &= \mM_{t-1} \odot \left( 1- \mE\vr_t \right) + \mA_t\\
		&= \left( \mM_{t-2} \odot \left( 1-\mE\vr_{t-1} \right) + \mA_{t-1} \right) \odot \left( 1 - \mE\vr_t \right) +\mA_t\\
		&= \sum_{k=0}^{\infty} \mA_{t-k} \odot \left( \overset{\odot}{\prod_{i=0}^{k-1}} \left( 1 - \mE\vr_{t-i} \right)  \right)\\
		&= \sum_{k=0}^{\infty} \mU_A^{T}\mH_{t-k} \mV_A \odot \left( \overset{\odot}{\prod_{i=0}^{k-1}} \left( 1-\mU_E^{T} \mH_{t-i}\mV_E \right)  \right)  
	\end{split}
\end{equation}
where $\overset{\odot}{\prod}$ signifies the indexed Hadamard product ( element-wise product). Hence, we see that with the addition of just one slot of memory $\mM_t$, the matrix read from it  $\mR_t$ would contain information about all the past states, which is a stark difference to the memory structure $\mQ_n$ also with only one slot of memory, where the only state read from $\mQ_n$ was $\mH_{t-2}$ (Eq. \ref{M-READ}). Since we avail the $\mR_t$ in \ref{NTM-access} to the state in the next time step, essentially we are giving each state access to all the past hidden states. This would imply the state solution at any time $t$ would contain infinitely many terms equivalent to the third term in Eq. \ref{mem-gen-sol-1}; one for each past state, thus making the asymptotic study of $\mM_t$, at least from a capacity standpoint, essentially infeasible.

Note that when there is more than one slot in $\mM_t$, then the weighing $\vw$ can select which past states to group together in a particular slot or more than one slot (Writing) and which group to retrieve (Reading), which would thus give it a type of \textit{selection} action over possibly each of the past states or the groups of past states stored in a slot which can be beneficial for the long-term recall tasks as shown by the experiments which follow.

\section{Simulation}
\label{S}

\begin{figure*}[htpb]
	\centering
	\includegraphics[width=\textwidth]{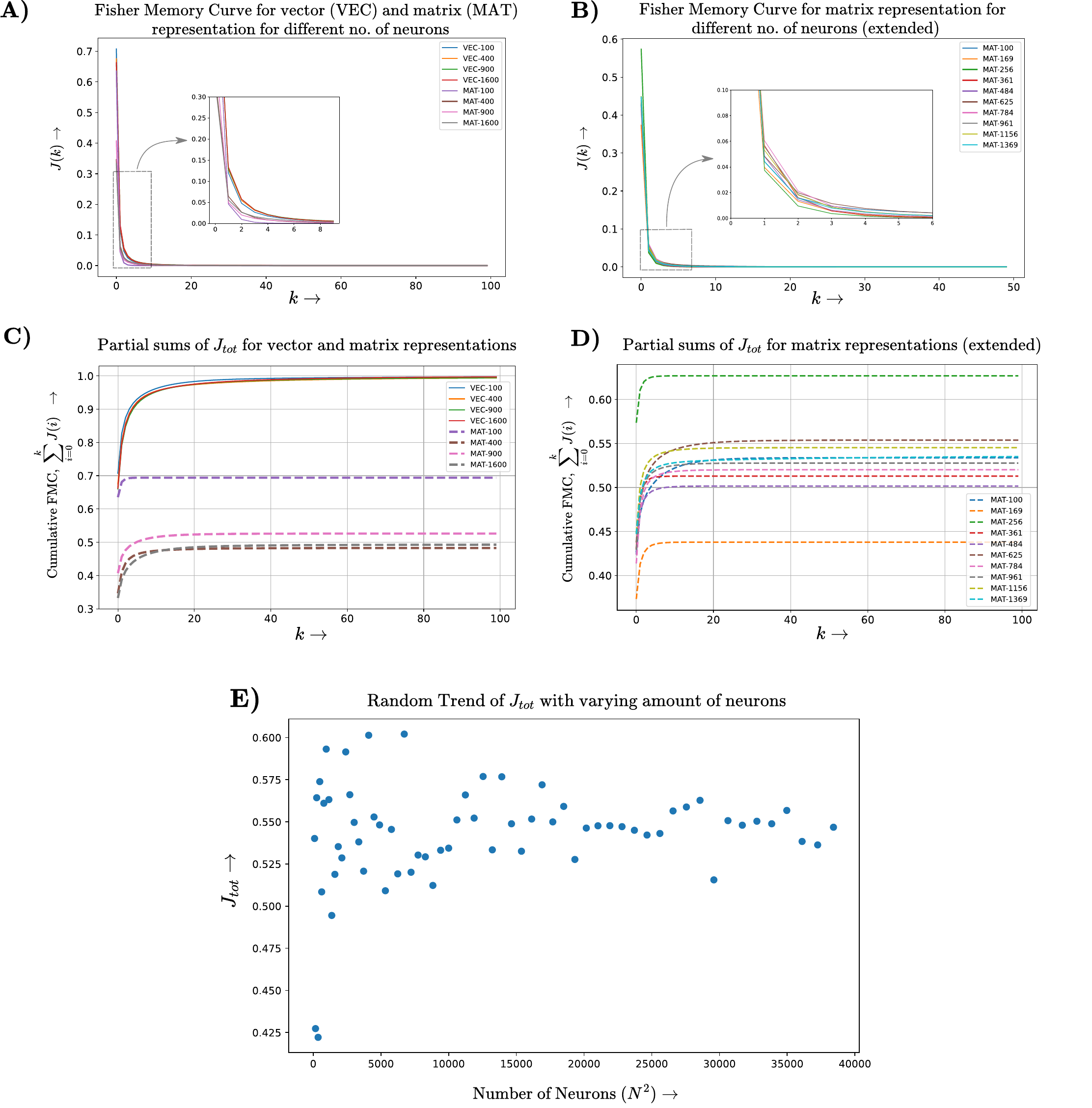}
	\caption{\textbf{A)} FMC for vector (Eq. \ref{fmm-vec-rep}) and matrix (Equation \ref{fc}) representation of networks with normal connections. \textbf{B)} FMC for matrix representation net for a larger number of neurons. \textbf{C)} Cumulative sum of $J(i)$ for vector and matrix RNNs. The sum asymptotically converges to 1 for vector nets while it remains $<1$ for matrix nets for any number of neurons as shown in C) and also in D) for a higher number of neurons. \textbf{E)} In the case of an even higher number of neurons for matrix nets, this figure shows their corresponding $J_{tot}$. It can be seen that the variance in capacity is small as the number of neurons is increased. All the above simulations assume $\Tr(\mW^T\mW) = 1$ and $\varepsilon_1 = \varepsilon_2 = 1$ so that $J_{{tot}_{rel}} = J_{tot}$.}
	\label{fig:simulationmat}
\end{figure*}
\begin{figure*}[htpb]
    \centering
    \includegraphics[width=\textwidth]{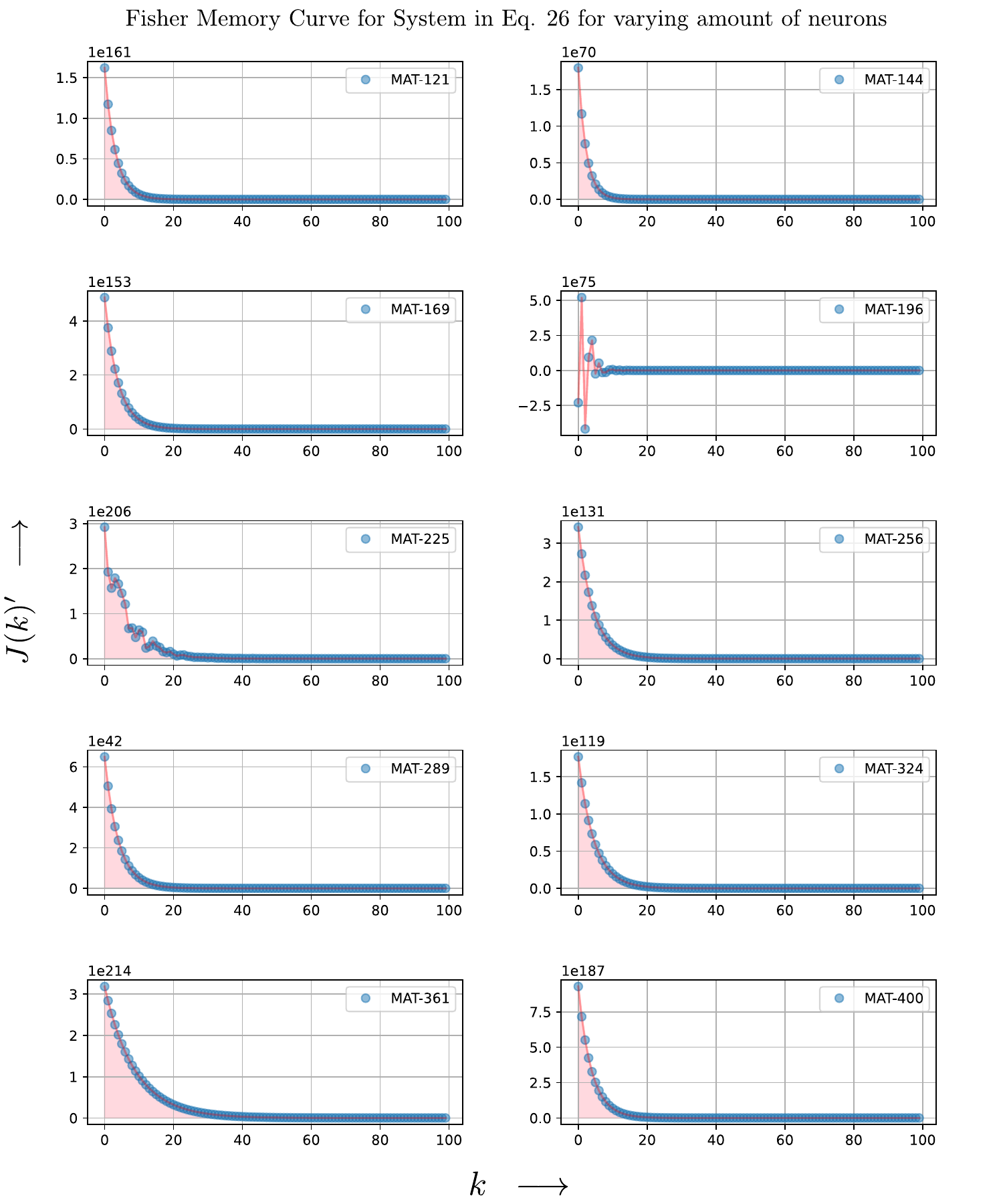}
    \caption{The FMC of the system in Eq. \ref{mem-dyn-sys} with $p=1$ for varying number of neurons. The area covered by each FMC, highlighted in pink, is clearly proportional to the memory capacity (Definition \ref{D-6} and see Figure \ref{fig:J_k_cumul_mem}). The apparent \textit{rough nature} of $J(k)^\prime$ for some FMCs as in MAT-196 \& MAT-225 might be due to the approximation of Eq. \ref{mean-cov-mem-dyn-sys} to a finite value of $m$ for practical realization. Note that we obtain the results for normal connections, as in general cases, frequent overflow can occur.} 
    \label{fig:J_k_prime_complete}
\end{figure*}
\begin{figure*}[htpb]
    \centering
    \includegraphics[width=\textwidth]{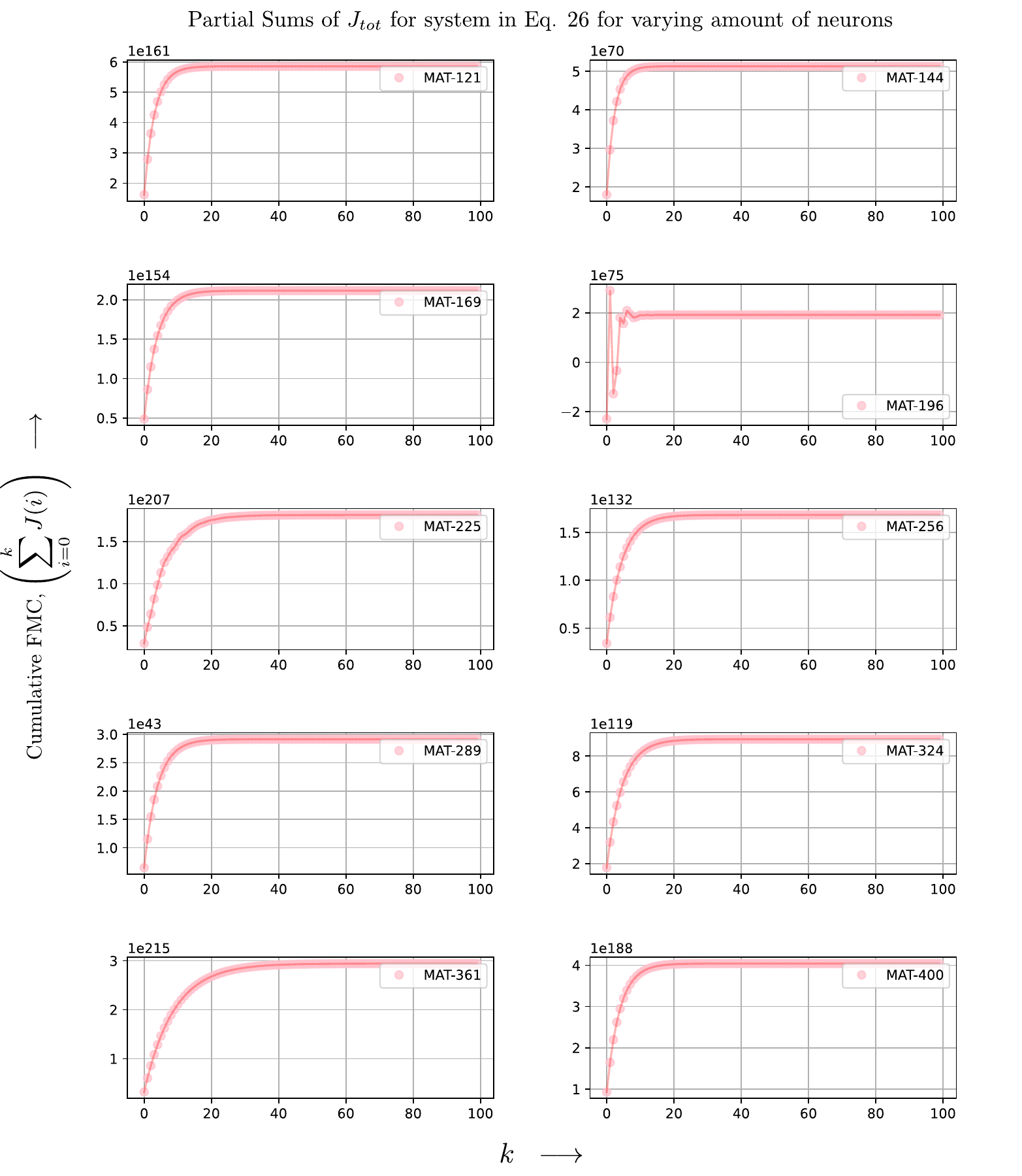}
    \caption{The cumulative sum of FMC at each time step for the system in Eq. \ref{mem-dyn-sys} with $p=1$ for a varying number of neurons, where we plot the FMC in Figure \ref{fig:J_k_prime_complete}. Since the connection matrices are normal and convergent, we see that even in such restrictive cases, the addition of even a single slot of memory blows up the memory capacity as the one compared to matrix representations without memory (Figure \ref{fig:simulationmat}, C \& D). There does not seem to be any relation between the capacities of RNNs with different numbers of neurons.} 
    \label{fig:J_k_cumul_mem}
\end{figure*}
 
Corollary \ref{C-10} and Theorem \ref{T-11} describe fundamental bounds on the memory capacity of certain types of matrix representation networks. However, it does not give much detail about how traces of memory would achieve this bound. Hence, we now present simulations of the capacity of systems in Eq. \ref{dyn-sys} and Eq. \ref{mem-dyn-sys} for normal connectivity and varying amounts of neurons. 

In particular, we see that the capacity $\mJ_{tot}$ of the linear matrix representation network with normal convergent connectivity matrices is always less than 1 (Theorem \ref{T-6}) whereas the capacity for vector representation networks sums to 1 as shown in Figure \ref{fig:simulationmat} C). The sharper decrease in the FMC for linear matrix nets can also be seen in Figure \ref{fig:simulationmat} A). We also see the apparent random nature of the memory capacity of such networks with varying number of neurons in Figure \ref{fig:simulationmat} E). Figure \ref{fig:J_k_prime_complete} shows the FMC and the capacity of the system \ref{mem-dyn-sys} with normal connectivity and varying number of neurons (note the overwhelming increase in the capacity).

 The main purpose of Figure 2 (A) is thus to compare the FMC of matrix and vector representation, and we clearly see that the curve for matrices (for any random number of neurons) goes to zero more quickly than the vector counterparts. Figure 2 (B) can be considered as an extension of Figure (A), just to portray that increasing neurons don’t affect FMC in any way for matrix networks. Since memory capacity is just the area under the FMC, Figure 2 (C) verifies Corollary 10 and Figure 2 (D) shows that there is no apparent relation between the relative memory capacity and the number of neurons.

One can compute $\mathbf{\Sigma}$ and $\mathbf{\Psi}$ using the discrete-time Lyapunov equations. Eq. \ref{mean-cov-state-sol} shows the infinite sum of the matrices needed to compute the noise covariances, note that they follow the discrete Lyapunov equations as given below
\begin{equation}
\label{dlyap}
    \begin{split}
        \mU^T\mathbf{\Psi} \mU + \mI &= \mathbf{\Psi}\\
        \mV^T\mathbf{\Sigma} \mV + \mI &= \mathbf{\Sigma}
    \end{split}
\end{equation}

However, for the memory capacity of the system in Eq. \ref{mem-dyn-sys} ($J_{tot}^\prime$) with $p=1$, we use the Eq. \ref{mean-cov-mem-dyn-sys}; where each of the terms in $\mathbf{\Sigma}_{\text{MEM}}$ and $\mathbf{\Psi}_{\text{MEM}}$ can be found by recursive use of discrete-time Lyapunov equation for $\mathbf{\Sigma}_{\text{STATE}}$ and $\mathbf{\Psi}_{\text{STATE}}$ (Eq. \ref{dlyap}), respectively. 

\section{Experiments}
\label{E}

In this section, we show the results of the MatNTM over two synthetic tasks while comparing it with a basic Matrix RNN. Since the addressing mechanism of the MatNTM is identical to  NTM, it is natural to expect the learned memory addressing scheme of MatNTM to be identical to that of NTM for the same type of tasks. However, since the main goal of the paper is to introduce higher-order input data to be processed directly, we are constrained to use matrix-shaped inputs instead of vectors. However,  since the usual NTM only accepts vector-shaped inputs, we construct a matrix analogue of each such vector task.

In order to keep the MatNTM closer to theoretic evaluations done previously, we perform minimal-to-no hyperparameter tuning. Only the size and number of hidden layers  
of Matrix RNN are changed in between tasks, which thus allows us to portray the difficulty faced by the MatNTM over various tasks much more explicitly. Note that the bilinear mapping between the states introduces quadratic form which can make the training unstable as can be seen from the learning curves in Figures \ref{fig:CopyLC} and \ref{fig:AssRecLC}. We use the RMSprop \cite{Tieleman2012} optimizer for all tasks with a learning rate of $10^{-4}$ with $\tanh$ non-linearity for all recurrent matrix layers. We noted that using any non-saturating non-linearities such as \verb|ReLU| and \verb|LeakyReLU| causes a frequent overflow in the operations of the addressing mechanisms of the MatNTM (Figure \ref{fig:matntm} -  Steps 1-4). We provide further details for the training parameters in Table \ref{tab:TrainDetails}.

\begin{table*}[htbp!]
\centering
\begin{tabular}{ |p{2cm}|M{1.5cm}|M{1.5cm}|M{1.5cm}|M{1.5cm}|M{1.5cm}|M{2cm}| }
 \hline
 Model & Batch Size & Input Shape & Hidden State &  Memory Size & Learning Rate & No. of Parameters  \\
 \hline
 MatNTM - Copy Task   & 16 & $[5,5]$ & $3\times [15,15]$ &  $[120,6,6]$ & $1\times 10^{-4}$ & 4121 \\
 MatNTM - Associative Recall Task & 16 & $[5,5]$ & $4\times [20,20]$ &  $[120,6,6]$ & $8\times 10^{-5}$ & 7946\\
 \hline
 Matrix RNN - Copy Task &   16 & $[5,5]$ & $3\times [15,15]$ & $-$ & $1\times 10^{-4}$ & 2175 \\
 Matrix RNN - Associative Recall Task & 16 & $[5,5]$ & $4\times [20,20]$ & $-$ & $8\times 10^{-5}$ & 5675 \\
 \hline
\end{tabular}
\label{tab:TrainDetails}
 \caption{Hyperparameters for models used for experiments. In the  Copy Task, we take $l$ to be random between 1 and 20, and in the Associative Recall Task, we fix $n=2$ and choose $k$ randomly between 2 and 10.}
\end{table*}

We present open source code of the implementation with results here \footnote{\url{https://github.com/sydney-machine-learning/Matrix_NeuralTuringMachine}}.

\subsection{Matrix Copy Task}

A long-standing problem in RNNs has been that of efficient recall of sequences observed over a long time duration in the past \cite{pmlr-v28-pascanu13} due to vanishing and exploding gradient problems arising during training. In order to benchmark the long-term memory capability of MatNTM, we test it on a version of the copy task as done in  \cite{hochreiter1997long,graves2014neural} which has been extended for matrix sequences. Note that this framework includes the vector copy task as a special case. 

Consider a sequence of $N\times N$ matrices denoted as $\{\mX_1,\mX_2,\dots,\mX_l\}$ where, for our experiments, $(X_i)_{jk} \sim \text{Bernoulli}\left(\frac{1}{2}\right)$ and $l$ can vary for each such sequence. Additionally, consider \textit{start-of-file ($\mX_{sof}$)} and \textit{end-of-file ($\mX_{eof}$)} delimiters which determine that the content that needs to be copied is present between them. These limiters are added to the sequence at the start and the end respectively to form the final sequence $\{\mX_n\} = \left\{ \mX_{sof}, \mX_1,\dots,\mX_l, \mX_{eof}\right\}$. Consider a model $L_\theta$ with parameters $\theta$, which takes   a set of matrix (two-dimensional) sequences as input, and gives out a fixed length matrix sequence of a predetermined shape

\[ \{\mY_n\} =  L_\theta\left( \{ \mX_n \} \right)\].

We train the model  on a copy task given the parameters $\theta$ are tuned so that output and input sequences are exactly the same in content and in order, i.e.
\[\{\mY_n\} = \{\mX_n\}\]
The training here is carried by minimizing the binary cross-entropy loss between input sequence $\{\mX_1,\dots,\mX_l\}$ and output sequences $\{\mY_1,\dots,
\mY_l\}$.  

\begin{figure*}[htbp!]
    \centering
    \includegraphics[width=\textwidth]{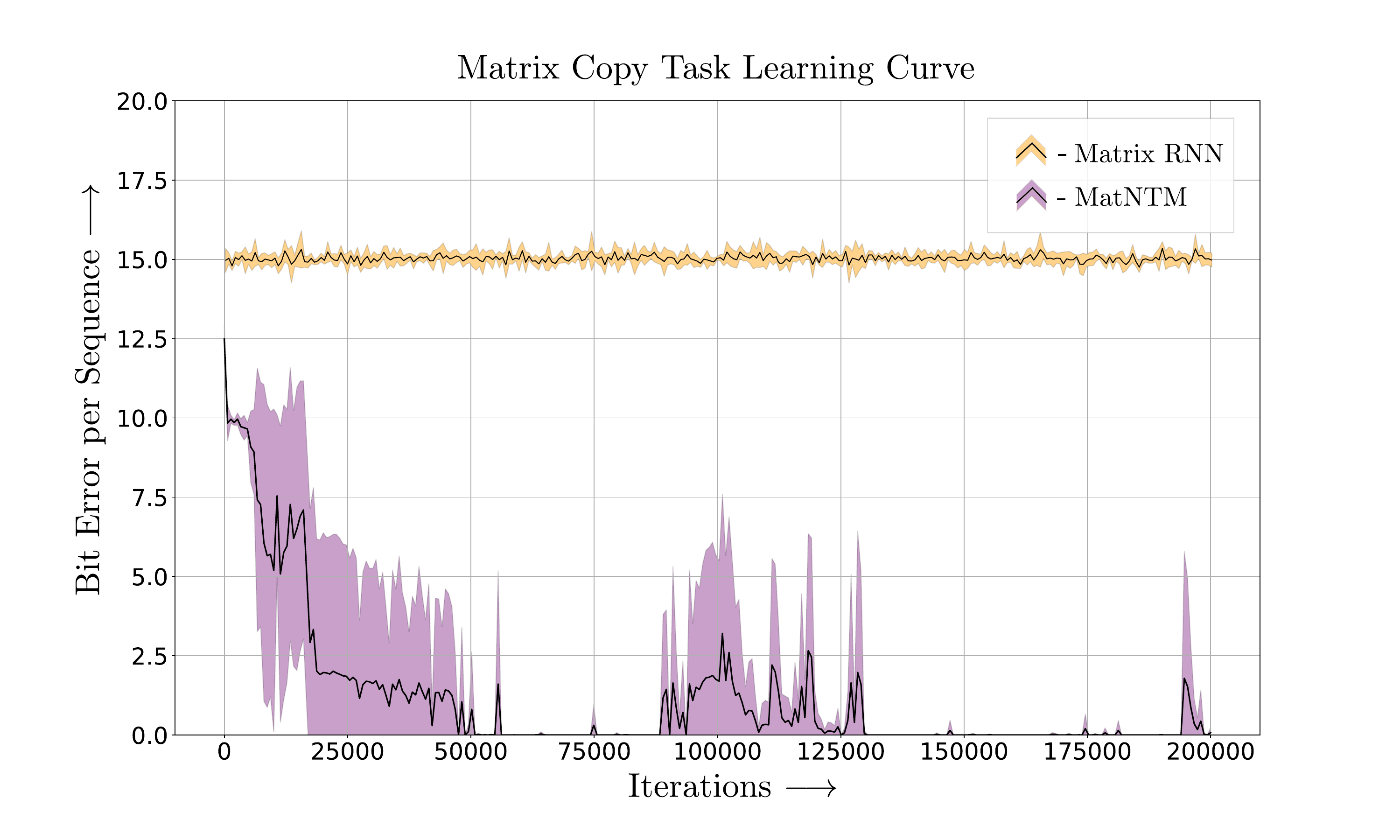}
    \caption{Matrix Copy Task learning curves for MatNTM and Matrix RNN averaged over five runs.}
    \label{fig:CopyLC}
\end{figure*}

Figure \ref{fig:CopyLC} shows that Matrix RNN fails to learn the copy task, whereas MatNTM learns this task since it fails to converge as the learning time increases.  The reason for this could be that the  Matrix RNN had a lack of precise memory of past input matrices and could not reconstruct the input sequence correctly. A simple addition of memory cells, as in the case of LSTM model would have been enough to learn this task for smaller length sequences; however, that would have required more parameters (weights and biases) to be trained. We show that the addition of an external memory makes this task learnable by the addition of very few more parameters as compared to LSTM models, i.e. as low as 5675. In contrast, the copy task (for vector sequences) in NTM with a feedforward controller takes close to 17,000 parameters \cite{graves2014neural}.
\begin{figure*}[htbp!]
    \centering
    \includegraphics[width=\textwidth]{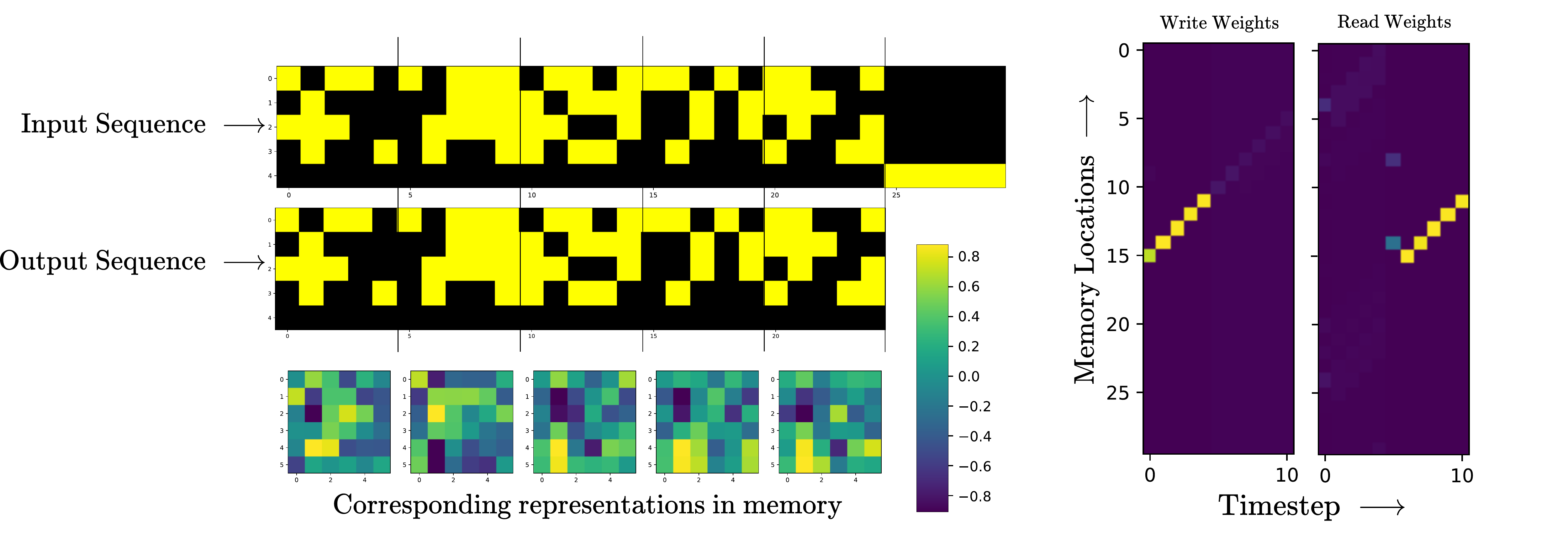}
    \caption{As expected, the MatNTM learns the same algorithm for a copy task similar to the NTM \cite{graves2014neural} while featuring matrix representation. Note that the input sequence contains an \textit{eof} delimiter channel on the last row and the last element of the input sequence acts as the actual \textit{eof} delimiter. The representation stored in memory corresponding to each input sequence is also shown. By incorporating matrix representations, we have essentially shortened the timespan required for processing the whole input sequence as a usual NTM would have required 30 time steps compared to just 6 in MatNTM, similarly for the output sequence. This is more evident in the right-hand side of the figure where we see the learned model actively iterating over the memory slots to write the corresponding representation while reading from the same locations in the original order.}
    \label{fig:CopyAblation}
\end{figure*}

\subsection{Matrix Associative Recall Task}

\begin{figure*}[htbp!]
    \centering
    \includegraphics[width=\textwidth]{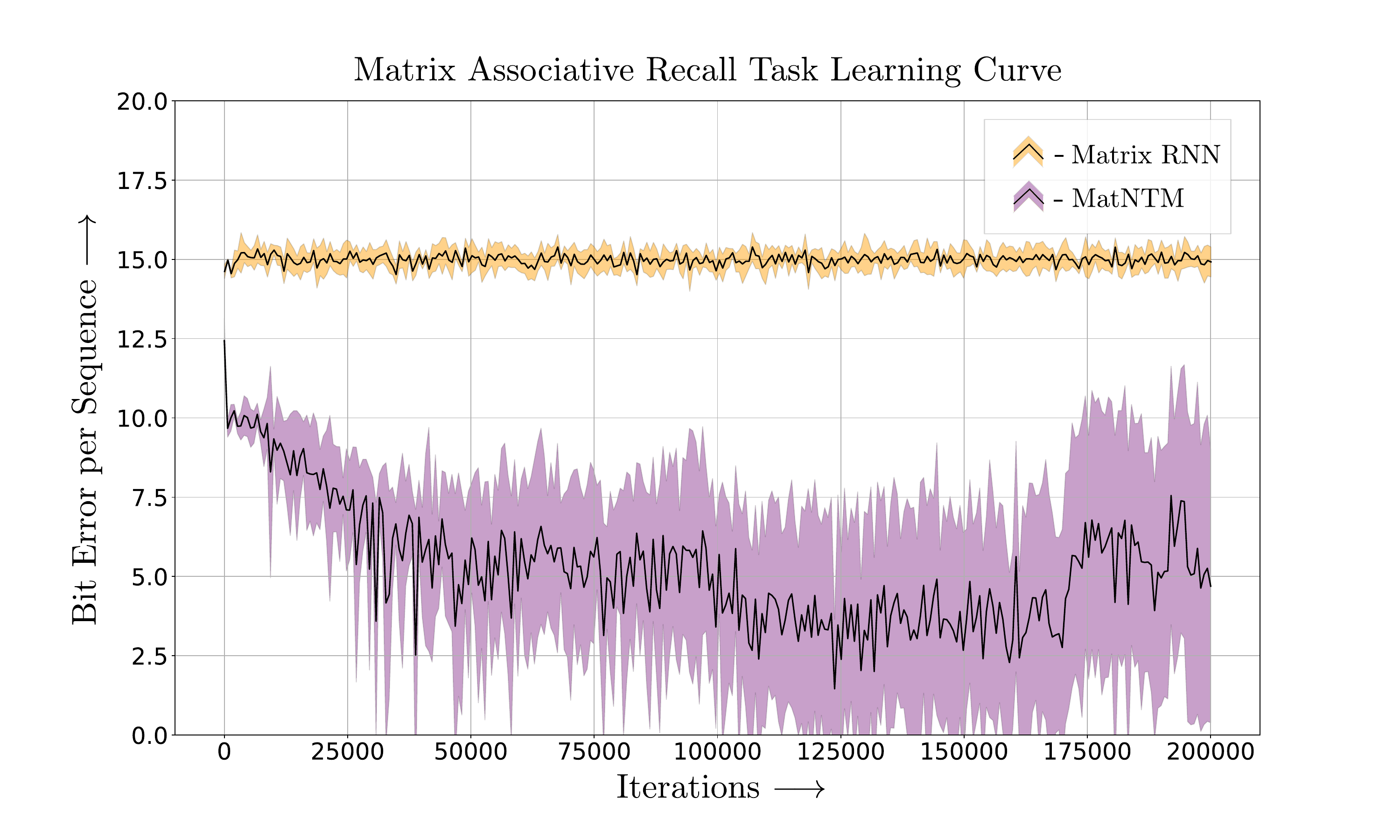}
    \caption{Matrix Associative Recall Task learning curves for MatNTM and Matrix RNN averaged over five experimental runs with different initialization in parameters, i.e. weights and biases.}
    \label{fig:AssRecLC}
\end{figure*}

The goal of this task is to test the model's capability to form links between the data it has previously seen. In particular, we form the following task, extending the one proposed in \cite{graves2014neural}.

Consider a sequence of matrices defined as an \textit{item} as $\mX_{\text{item}_i} = \{\mX_{sof},\mX_{1_i},\dots,\mX_{n_i},\mX_{eof}\}$ for $i = 1,\dots, k$, where $n$ is kept fixed while training whereas the number of items $k$ can vary. Now, the complete sequence of inputs shown to the network can be written as $\{ \mX_{\text{item}_1},\dots,\mX_{\text{item}_k}\}$. After processing this input, the network is shown a special end of input delimiter $\mX_{\text{delim}}$ which signifies the end of input sequence and the beginning of a query which is $\mX_{\text{query}} = \{ \mX_{1_c},\dots,\mX_{n_c} \}$, where $c\sim \text{Unif}(1,2,\dots,k-1)$. The target in this task is to thus output the next item in the input sequence processed earlier, i.e. for the model $L_\theta$, the output sequence $\{\mY_n\} = L_\theta\left( \{\mX_n\} \right)$ should be,
\[ \{\mY_n\} = \{ \mX_{1_{c+1}},\dots,\mX_{n_{c+1}} \} .\]
This task hence evaluates the learner's capability to form an association between the target and the query based on the past sequences it has been provided.

\begin{figure*}[htbp!]
    \centering
    \includegraphics[width=\textwidth]{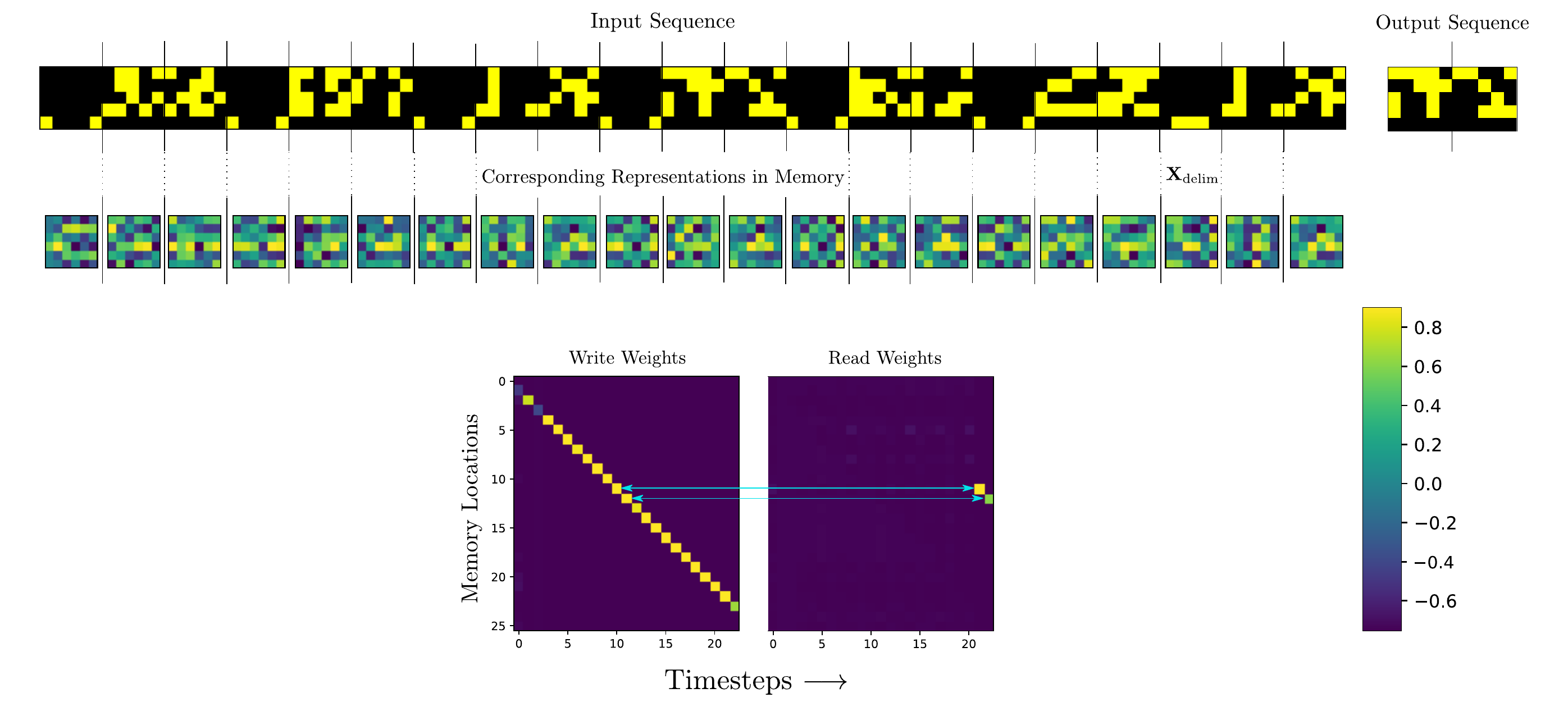}
    \caption{The top-most sequence denotes the input given to MatNTM and the output sequence received from the MatNTM. The next row shows the matrices stored in memory corresponding to that input (and combined with the information of all the cumulative inputs provided by controller RNN). Moreover, the bottom row shows the Read \& Write weights distributed temporally which shows clearly the shift in read head to the memory location of the matrix next to the query. The solution learned for this task, as suggested by the bottom row  is identical to the results in the literature \cite{graves2014neural}.}
    \label{fig:AssRecAblation}
\end{figure*}

We can clearly see in Figure \ref{fig:AssRecLC} that matrix RNNs cannot learn this task as well, whereas MatNTM seems to work better; however, it is still not learning completely (except on 2 runs out of 5 where the model gave zero error). Figure \ref{fig:AssRecAblation} further depicts that the algorithm learned by MatNTM is identical to usual NTMs, though with matrix representations.
\begin{figure*}
    \centering
    \includegraphics[width=\textwidth]{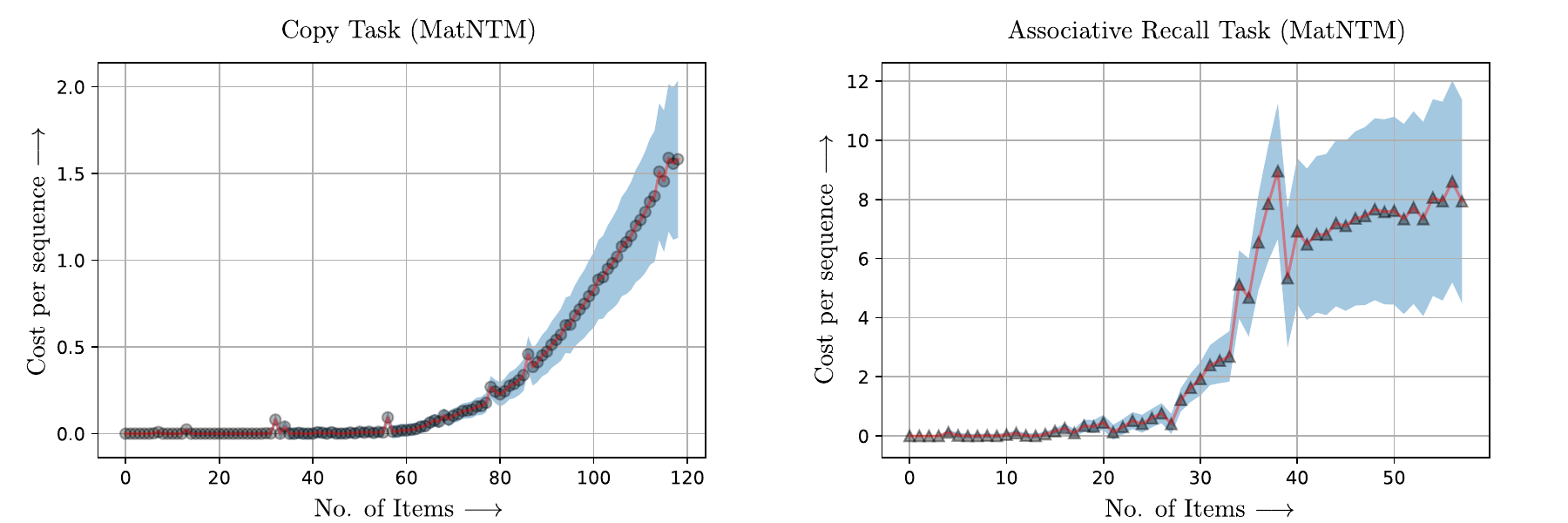}
    \caption{Results for trained models of MatNTM on respective tasks. The left figure shows the increase in cost with an increase in the length $l$ of the sequence of matrices in the copy task. The right figure shows the increase in cost with an increase in the length of an item $n$. }
    \label{fig:BSE}
\end{figure*}

\section{Discussion and Future Work}\label{Sec-DaFW}

We studied the memory capacity of a new class of RNNs introduced by Gao et. al \cite{gao2016matrix} for incorporating matrix representations using a bilinear map. We discussed various existing definitions of memory capacity and how most of the definitions prove to be not easily generalizable to matrix representations. We hence use a probabilistic model of the memory capacity using Fisher information as introduced by Ganguli et. al \cite{Ganguli18970}. We investigated how the memory capacity for matrix representation networks is limited under various constraints, and in general, without any constraints. In the case of memory capacity without any constraints, we found that the upper bound on memory capacity is $N^2$ for $N\times N$ state matrix. This seems to generalize the similar upper bound for vector representation RNNs ($N$ for state vector of length $N$). Moreover, we demonstrated that the inclusion of saturating non-linearities over the state transition may further bind the total memory capacity even tighter.

\textcolor{black}{We note that the proposed method is based on a  Gaussian assumption. The main equation  (Eq. 5) is based solely on the assumption that the noise channel is a Gaussian matrix with zero mean matrix and covariance matrices (for rows and columns) which are diagonal with all eigenvalues equal (introduction of Sec. 3). The main reason of taking Gaussian distribution is due to the literature (Ganguli et. al. \cite{Ganguli18970}) which uses the Gaussian assumption on the noise.  Corollary 10 shows that the relative memory capacity of Eq. 5 is lower than 1 which is based on the same assumptions as that of Ganguli et. al \cite{Ganguli18970} which used the vector representation rather than matrix representation. The main theorem from which our analysis of Section 3 begins would not be possible without the Gaussian assumption. The FMM of Eq. 5, as calculated in Theorem 2, would not have been possible without the Gaussian distribution. Even in that case, we had to compute the KL-divergence of a Gaussian random matrix distribution (Lemma 3) and as the proof shows, it is not at all clear how we can do the same for any other   distribution.}

In Figures 5 and 7, we find that the values of MatNTM fluctuate more largely than those of Matrix RNNs. Although we don’t have a definite answer to explain the main reason behind the difference, some of the reasons could be as follows.  It was our general observation that we had to take an unusually small learning rate during backpropagation for MatNTM in order to avoid not a number (NaN). We had to fine-tune our settings too much, for we saw that under certain random seeds, we were not able to converge to a minimum. Even then, as is visible in Figure 5 and 7, the loss fluctuates a lot. This indicates that the loss landscape is quite chaotic for matrix representations with memory and is very sensitive to starting conditions. This is to be expected since Eq. 5 introduces a quadratic form in the recurrent system (as compared to vector representation networks) which is expected to make the loss landscape quite a lot more sensitive to fine differences in initial conditions. Memory networks in general have a rather chaotic loss curve due to the increase in complexity of the model that comes from having a differentiable memory which is visible in the literature  (Graves et. al  \cite{graves2016hybrid}).

The theoretical study and subsequent simulations hence reveal the fundamental bounds on memory capacity as revealed by the proposed definition and the striking increase in memory capacity induced by external memory, which is in line with basic intuition, but quantitatively has been a difficult task to ascertain. Moreover, we used a notion of memory capacity based only on Fisher Information, whereas much of the recent work is focused on functional \& probabilistic definitions  \cite{vershynin2020memory}. Counter-acting the difficulty of extending such notions to the domain of matrix representations would be the obvious next step.

One of the recent developments in the class of neural network architectures has been memory-augmented neural networks. However, quantifying the increase in actual memory capacity induced by such an external memory hasn't been discussed yet. Given this motivation, we derived the memory trace of a linear matrix recurrent network with queue-like state memory, which provides the current state with the information of not just the state at $t-1$ timestep, but states at $t-p$ timesteps for a fixed $p\ge1$. We note that this is very similar to the work of Soltani and Jiang \cite{soltani2016higher} dubbed as Higher Order RNNs, which also provide the current state with a fixed number of past states, though in vector-based representation.

 In terms of mathematical limitations, we note that the focus of our study was a linearized version of Eq. 5 which is not suited to understand the general behavior of such networks under non-linearities.  The notion of memory capacity used in our study is not explicit, and in literature, there are various other notions of memory capacity that include analytic definitions as well, which in some sense is more explicit (see Section 2). We chose to work with a probabilistic definition, as we were able to develop the tools required for the analysis of matrix representations under such a definition in a rather straightforward manner.   We along with others who studied matrix neural networks \cite{gao2017matrix,8489077}  argue that it prevents the loss of spatial coherence introduced by vectorizing two-dimensional data. Unfortunately, as of now, we don’t have any way of proving/disproving this mathematically. The major experimental limitation we faced was the extreme sensitivity of MatNTM to starting conditions and hyperparameters. This greatly inhibits our potential to carry out a diverse set of experiments on MatNTM, as is done by others in the field of memory networks. We note that memory networks such as  NTMs already are quite difficult to train, and the addition of more complicated representations makes the training process much harder.

\textcolor{black}{There is a high increase in the complexity of our model by the introduction of matrix representation when compared to conventional vector representation. In particular, for Eq. 5, we see that one can view that matrix recurrent system as a stacked version of feedforward neural networks, one for rows of $X$ and the other for columns of $X$. Let us explain. The main part of the equation that affects the state is $X(n) = U^{TX(n-1)V}$. One can view $U^{TX(n-1)}$ as a simple neural network with a weight matrix $U^T$ which acts on each column of $X(n-1)$. The effect of V as in $U^{TX(n-1)}$, $V$ is simply as another feed-forward network with weight matrix $V^T$ which acts on the rows of the resultant matrix $U^{TX(n-1)}$. It is in this manner that matrix networks can be thought of as classical feed-forward networks stacked both row-wise and column-wise. It clearly appears that even a simple matrix network has more complexity than its vector-based simple neural network counterpart. Adding to this external memory, one can expect that the complexity of the model increases substantially. Another way of looking at this could be as follows. In Table 1, we see that both Matrix RNNs and MatNTMs do not require too many parameters (compare this to vector counterparts, where NTM with feedforward controller requires 17162 parameters to learn basic copy task  \cite{graves2014neural,graves2016hybrid}. This indicates that, in particular, MatNTM learns the same task in a much smaller number of parameters than its vector counterpart (NTM). This is only possible when the complexity of the model itself is such that it allows the model training even under such constraints posed by a low number of parameters. Hence, one can conclude that the complexity of any model which is learning a task with fewer parameters than some other one is bound to be more complicated. But it is important to keep in mind that this is not for free as we suffer greatly while finding the right hyperparameters to train a complex model such as MatNTM.}

Extending the memory networks to different types of representations has been proposed in recent times \cite{pham2018graph, khasahmadi2020memory}. We extend this line of work by extending the NTM with matrix representations whose preliminary results on two synthetic tasks show its advantage over the simple Matrix RNN. Due to our theory for the memory of matrix networks in Sections (\ref{MCiNN}-\ref{TEoESMoSD}), it was developed only with basic recurrent dynamics in mind, not for the widely used LSTMs due to obvious difficulties for the first study of such kind; hence, we refrained from using LSTM controller in MatNTM for the experiments. However, our experiments revealed that MatNTM with Matrix LSTM performs far better than that of the MatNTM with Matrix RNN controller presented in this work. We hope to extend this study to Matrix LSTM models in the future. It is usually the case that memory networks are trained on algorithmic tasks as given in earlier works (\cite{graves2014neural,graves2016hybrid}). Furthermore,   there is scope for the extension of  Matrix RNN to explore its potential for prediction and decision-making using time series data.

Our work introduced matrix representations as a natural next step from the vector representations as envisioned earlier \cite{gao2016matrix} and subsequently expanded   \cite{do2017learning}. Several such extensions have been made ever since, which go one step further and generalize
the notion of neural networks to higher-order tensors as the inputs. Clearly, such extensions are heavily non-trivial to work with from any, functional or probabilistic, point of view. Perhaps that is where an information geometric point of view might be helpful for any future work in this line of study.

Another way ahead is to develop a robust uncertainty quantification framework via Bayesian inference for Bayesian Matrix  NTMs. We note that the copy tasks by NTMs are computationally expensive and Bayesian inference via Markov Chain Monte Carlo Methods (MCMC) requires thousands of samples or model realizations for sampling the posterior distribution of neural weights and biases.  We can incorporate recent frameworks that used MCMC with gradient methods and parallel computing \cite{CHANDRA2019NC} to overcome computational challenges. Moreover, surrogate-based MCMC methods for computationally expensive models can also be used \cite{CHANDRA2020EAI} along with variational inference methods \cite{blundell2015weight}.

\section*{Data and Code Availability }

We provide data and code via GitHub repository \footnote{\url{https://github.com/sydney-machine-learning/Matrix\_NeuralTuringMachine}}

\newpage
\newpage

\newpage
\newpage
\appendix
\section{Proof of Lemma \ref{L-3}}
\label{Appendix-A1}
Consider two normally distributed random matrices $\rmX_1 \sim \mathcal{MN}_{n\times p}\left( \mM_1,\mA_1,\mB_1 \right) $ and $\rmX_2 \sim \mathcal{MN}_{n\times p}\left( \mM_2,\mA_2,\mB_2 \right) $. We have that
\begin{equation*}
	\begin{split}
		&p\left( \rmX_i \right)=\\& (2\pi)^{-\frac{1}{2}np}\abs{\mA_i}^{-\frac{1}{2}p}\abs{\mB_i}^{-\frac{1}{2}n} \exp\left[ -\frac{1}{2}\Tr\left( \mB_i^{-1}\left( \mX - \mM_i \right) ^{T}\mA_i^{-1}\left( \mX - \mM_i \right) \right)  \right] 
	\end{split}
\end{equation*}
for both $i \in \left\{ 1,2 \right\} $.
\newline\newline
The KL-divergence between two distributions $p_1$ and $p_2$ is given by (note $\log$ here is the natural logarithm),
\begin{equation*}
	\begin{split}
		\KL\left( p_1 \Vert p_2 \right) = \mathbb{E}_{p_1}\left[ \log \frac{p_1}{p_2} \right] .
	\end{split}
\end{equation*}
Hence, the KL-divergence between $p\left( \rmX_1 \right) $ and $p\left( \rmX_2 \right) $ can be calculated as,
\begin{equation*}
	\begin{split}
		\KL\left( p\left( \rmX_1 \right) \Vert p\left( \rmX_2 \right)  \right) &= \mathbb{E}_{p\left( \rmX_1 \right) } \left[ \log p\left( \rmX_1 \right) - \log p\left( \rmX_2 \right)  \right] \\
		&= \underbrace{\mathbb{E}_{p\left( \rmX_1 \right) }\left[ \log p\left( \rmX_1 \right)  \right]}_{\text{PART 1}} - \underbrace{\mathbb{E}_{p\left( \rmX_1 \right) }\left[ \log p\left( \rmX_2 \right)  \right]}_{\text{PART 2}}.
		\end{split}
\end{equation*}
We first calculate PART 2 as follows,
\begin{equation*}
	\begin{split}
		&\mathbb{E}_{p\left( \rmX_1 \right)}\left[ \log p\left( \rmX_2 \right)  \right]\\ &=  \mathbb{E}_{p\left( \rmX_1 \right)} \Biggl[ -\underbrace{\log\left( \left( 2\pi \right) ^{\frac{1}{2}np} \abs{\mA_2}^{\frac{1}{2}p} \abs{\mB_2}^{\frac{1}{2}n} \right)}_{\text{constant : }C_2}\\ &\;\;\;\;\;\;\;\;\;\;\;\;\;-\frac{1}{2}\Tr\left( \mB_2^{-1} \left( \mX - \mM_2 \right) ^{T}\mA_2^{-1}\left( \mX - \mM_2 \right)  \right)  \Biggl]\\
		&= -C_2 -\frac{1}{2} \mathbb{E}_{p\left( \rmX_1 \right) } \left[ \Tr\left( \left( \mX - \mM_2 \right)^{T}\mA_2^{-1}\left( \mX-\mM_2 \right) \mB_2^{-1} \right)  \right] \\
		&= -C_2 - \frac{1}{2} \mathbb{E}_{p\left( \rmX_1 \right) }\left[ \sum_{i=1}^{p} \sum_{j=1}^{p} \sum_{k=1}^{n} \sum_{t=1}^{n} \left( x_{ti} - m_{2;ti} \right) \left( x_{kj} - m_{2;kj} \right) u_{2;tk}^{-1} v_{2;ji}^{-1} \right] \\
		&= -C_2 - \frac{1}{2} \mathbb{E}_{p\left( \rmX_1 \right) }\Biggl[ \sum_{i=1}^{p} \sum_{j=1}^{p} \sum_{k=1}^{n} \sum_{t=1}^{n}  u^{-1}_{2;tk}v^{-1}_{2;ji}\Biggl( x_{ti}x_{kj} - m_{2;ti} x_{kj}\\ &\;\;\;\;\;\;\;\;\;\;\;\;\;\;\;\;\;\;\;\;\;\;\;\;\;\;\;\;\;\;\;\;\;\;\;\;\;\;\;\;\;\;\;\;\;\;\;\;\;\;\;\;\;\;\;\;\;\;\;\;-m_{2;kj}x_{ti} + m_{2;ti}m_{2;kj}\Biggl)\Biggl]
	\end{split}
\end{equation*}
Now, using the fact that $\mathbb{E}\left[ x_{i_1 j_1}x_{i_2 j_2} \right] = a_{i_1 i_2}b_{j_1 j_2} + m_{i_1 j_1}m_{i_2 j_2}$ and $\mathbb{E}\left[ x_{i_1 j_1} \right] = m_{i_1 j_1}$ \textcolor{black}{(Theorem \ref{T-6}.3.3,  \cite{gupta1999matrix})} after taking expectation inside the sum, we get:
\begin{equation*}
	\begin{split}
		&\mathbb{E}_{p\left( \rmX_1 \right)}\left[ \log p\left( \rmX_2 \right)  \right]\\ &= -C_2 -\frac{1}{2}  \sum_{i=1}^{p} \sum_{j=1}^{p} \sum_{k=1}^{n} \sum_{t=1}^{n} u^{-1}_{2;tk}v^{-1}_{2;ji} a_{1;tk}b_{1;ij}\\&\;\;\;\;\;\;\;\;\;\;\;\;\;\;\;\;\;\;\;\;\;\;\;\;\;\;\;\;\;\;\;\;\;\;\;\; + m_{1;ti}m_{1;kj}u^{-1}_{2;tk}v^{-1}_{2;ji}\\ &\;\;\;\;\;\;\;\;\;\;\;\;\;\;\;\;\;\;\;\;\;\;\;\;\;\;\;\;\;\;\;\;\;\;\;\; -  u^{-1}_{2;tk}v^{-1}_{2;ji}m_{2;ti} m_{1;kj} \\
		&\;\;\;\;\;\;\;\;\;\;\;\;\;\;\;\;\;\;\;\;\;\;\;\;\;\;\;\;\;\;\;\;\;\;\;\; - u^{-1}_{2;tk}v^{-1}_{2;ji} m_{2;kj}m_{1;ti}\\
		&\;\;\;\;\;\;\;\;\;\;\;\;\;\;\;\;\;\;\;\;\;\;\;\;\;\;\;\;\;\;\;\;\;\;\;\; +  u^{-1}_{2;tk}v^{-1}_{2;ji} m_{2;ti}m_{2;kj}\\
		&=-C_2 -\frac{1}{2}  \sum_{i=1}^{p} \sum_{j=1}^{p} \sum_{k=1}^{n} \sum_{t=1}^{n}  u^{-1}_{2;tk}v^{-1}_{2;ji} a_{1;tk}b_{1;ij}\\
		&\;\;\;\;\;\;\;\;\;\;\;\;\;\;\;\;\;\;\;\;\;\;\;\;\;\;\;\;\;\;\;\;\;\;\;\; + v^{-1}_{2;ji}m_{1;kj}u^{-1}_{2;tk}m_{1;ti}\\
		&\;\;\;\;\;\;\;\;\;\;\;\;\;\;\;\;\;\;\;\;\;\;\;\;\;\;\;\;\;\;\;\;\;\;\;\; - v^{-1}_{2;ji}m_{1;kj}u^{-1}_{2;tk}m_{2;ti}\\
		&\;\;\;\;\;\;\;\;\;\;\;\;\;\;\;\;\;\;\;\;\;\;\;\;\;\;\;\;\;\;\;\;\;\;\;\; -v^{-1}_{2;ji} m_{2;kj}u^{-1}_{2;tk}m_{1;ti}\\
		&\;\;\;\;\;\;\;\;\;\;\;\;\;\;\;\;\;\;\;\;\;\;\;\;\;\;\;\;\;\;\;\;\;\;\;\; + v^{-1}_{2;ji} m_{2;kj}u^{-1}_{2;tk} m_{2;ti}\\
		&= -C_2 - \frac{1}{2} \Bigl(  \Tr( \mB_2^{-1}\mM_1^{T}\mA_2^{-1}\mM_1-\mB_2^{-1}\mM_1^{T}\mA_2^{-1}\mM_2\\&\;\;\;\;\;\;\;\;\;\;\;\;\;\;\;\;\;\;- \mB_2^{-1}\mM^{T}_2\mA_2^{-1}\mM_1  +  \mB_2^{-1}\mM_2^{T}\mA_2^{-1}\mM_2)\\&\;\;\;\;\;\;\;\;\;\;\;\;\;\;\;\;\;\; +\Tr\left( \mA^{-1}_2\mA_1 \right) \Tr\left( \mB_2^{-1}\mB_1 \right)\Bigl).
	\end{split}
\end{equation*}
Note that we interchanged the order of summation of traces in the last line only for the purpose of better representation. After some more rearrangement of terms inside the first trace, we get the following solution of PART 2,
\begin{equation*}
	\begin{split}
		&\mathbb{E}_{p\left( \rmX_1 \right)}\left[ \log p\left( \rmX_2 \right)  \right]\\ &= -C_2 - \frac{1}{2}\Tr\left( \mA^{-1}_2\mA_1 \right) \Tr\left( \mB_2^{-1}\mB_1 \right)\\&\;\;\; - \frac{1}{2}\Tr\left( \mB_2^{-1} \left( \mM_2-\mM_1 \right)^{T}\mA_2^{-1}\left( \mM_2-\mM_1 \right)  \right) 
	\end{split}
\end{equation*}
Calculating PART 1 in a similar path yields us the following,
\begin{equation*}
	\begin{split}
		&\mathbb{E}_{p\left( \rmX_1 \right) } \left[ \log p\left( \rmX_1 \right)  \right]\\ &=  \mathbb{E}_{p\left( \rmX_1 \right)} \Biggl[ -\underbrace{\log\left( \left( 2\pi \right) ^{\frac{1}{2}np} \abs{\mA_1}^{\frac{1}{2}p} \abs{\mB_1}^{\frac{1}{2}n} \right)}_{\text{constant : }C_1}\\
		&\;\;\;\;\;\;\;\;\;\;\;\;\;-\frac{1}{2}\Tr\left( \mB_1^{-1} \left( \mX - \mM_1 \right) ^{T}\mA_1^{-1}\left( \mX - \mM_1 \right)  \right)  \Biggl]\\
	&= -C_1 - \frac{1}{2} \mathbb{E}_{p\left( \rmX_1 \right) }\Biggl[ \sum_{i=1}^{p} \sum_{j=1}^{p} \sum_{k=1}^{n} \sum_{t=1}^{n}  u^{-1}_{1;tk}v^{-1}_{1;ji}\Biggl( x_{ti}x_{kj} - m_{1;ti} x_{kj}\\&\;\;\;\;\;\;\;\;\;\;\;\;\;\;\;\;\;\;\;\;\;\;\;\;\;\;\;\;\;\;\;\;\;\;\;\;\;\;\;\;\;\;\;\;\;\;\;\; -m_{1;kj}x_{ti} + m_{1;ti}m_{1;kj}\Biggl)\Biggl].
	\end{split}
\end{equation*}
Using the same two theorems used in calculations for PART 2, we get the following solution of PART 1,
\begin{equation*}
	\begin{split}
			&\mathbb{E}_{p\left( \rmX_1 \right) } \left[ \log p\left( \rmX_1 \right)  \right]\\ &= -C_1 - \frac{1}{2} \sum_{i=1}^{p} \sum_{j=1}^{p} \sum_{k=1}^{n} \sum_{t=1}^{n}  u^{-1}_{1;tk}v^{-1}_{1;ji}\Bigl( a_{1;tk}b_{1;ij} + m_{1;ti} m_{1;kj}  - m_{1;ti} m_{1;kj}\\
		&\;\;\;\;\;\;\;\;\;\;\;\;\;\;\;\;\;\;\;\;\;\;\;\;\;\;\;\;\;\;\;\;\;\;\;\;\;\;\;\;\;\;\;\;\;\;\;\;\;\; -m_{1;kj}m_{1;ti} + m_{1;ti}m_{1;kj}\Bigl)\\
		&=  -C_1 - \frac{1}{2} \sum_{i=1}^{p} \sum_{j=1}^{p} \sum_{k=1}^{n} \sum_{t=1}^{n}  u^{-1}_{1;tk}v^{-1}_{1;ji} a_{1;tk}b_{1;ij}\\
		&= -C_1 - \frac{1}{2}\Tr\left( \mA_1^{-1}\mA_1 \right) \Tr\left( \mB_1^{-1}\mB_1 \right)\\
		&= -C_1 -\frac{1}{2}np
	\end{split}
\end{equation*}
Thus, combining the solutions of both PART 1 and PART 2 yields us,
\begin{equation*}
	\begin{split}
		&\KL\left( p\left( \rmX_1 \right) \Vert p\left( \rmX_2 \right) \right)\\& = -C_1 - \frac{1}{2}np + C_2 + \frac{1}{2}\Tr\left( \mA^{-1}_2\mA_1 \right) \Tr\left( \mB_2^{-1}\mB_1 \right)\\
		&\;\;\;\;+ \frac{1}{2}\Tr\left( \mB_2^{-1} \left( \mM_2-\mM_1 \right)^{T}\mA_2^{-1}\left( \mM_2-\mM_1 \right)  \right)
	\end{split}
\end{equation*}
Substituting the values of $C_1$ and $C_2$ back in their respective places, we finally get the required form,
\begin{equation*}
	\begin{split}
		&\KL\left( p\left( \rmX_1 \right) \Vert p\left( \rmX_2 \right) \right)\\&= \log \frac{\abs{\mA_2}^{\frac{1}{2}p}\abs{\mB_2}^{\frac{1}{2}n}}{\abs{\mA_1}^{\frac{1}{2}p}\abs{\mB_1}^{\frac{1}{2}n}}  - \frac{1}{2}np + \frac{1}{2}\Tr\left( \mA^{-1}_2\mA_1 \right) \Tr\left( \mB_2^{-1}\mB_1 \right)\\
		&\;\;\;\;+ \frac{1}{2}\Tr\left( \mB_2^{-1} \left( \mM_2-\mM_1 \right)^{T}\mA_2^{-1}\left( \mM_2-\mM_1 \right)  \right)
	\end{split}
\end{equation*}
This proves Lemma \ref{L-3}.\qed




\newpage

\bibliographystyle{elsarticle-num} 

\bibliography{sample,2016}

\end{document}